\documentclass{article}

\usepackage{arxiv}

\usepackage[utf8]{inputenc}
\usepackage[T1]{fontenc}
\usepackage{hyperref}
\usepackage{url}
\usepackage{booktabs}
\usepackage{amsmath,amsfonts,amssymb}
\usepackage{nicefrac}
\usepackage{microtype}
\usepackage{graphicx}
\usepackage[numbers,round,sort&compress]{natbib}
\usepackage{doi}
\usepackage{bm}
\usepackage{xcolor}
\usepackage{array}
\usepackage{float}
\usepackage[labelfont=bf,labelsep=period,figurename=Fig.]{caption}
\graphicspath{{figures/}}
\hypersetup{colorlinks=true, allcolors=blue}
\expandafter\def\expandafter\UrlBreaks\expandafter{\UrlBreaks\do\-}

\makeatletter
\newcommand{\runinperiod}[1]{#1.}
\renewcommand{\subsubsection}{%
  \@startsection{subsubsection}{3}{\z@}%
                {1.5ex \@plus 0.5ex \@minus 0.2ex}%
                {-0.6em}%
                {\normalsize\bfseries\itshape\runinperiod}%
}
\makeatother
\usepackage{subfiles}
\title{AI writing on social media is associated with greater engagement and productivity}

\author{
  Changyang Wu\textsuperscript{a}, Lynnette Hui Xian Ng\textsuperscript{a}, Meryl Ye\textsuperscript{a}, Yutong Zhang\textsuperscript{b}, \\
  \bfseries James K. He\textsuperscript{c}, Felix Wallis\textsuperscript{c}, and Steve Rathje\textsuperscript{d,}\thanks{To whom correspondence should be addressed. E-mail: srathje@andrew.cmu.edu} \\[1.5ex]
  \textsuperscript{a}Software and Societal Systems Department, Carnegie Mellon University, Pittsburgh, PA 15213 \\
  \textsuperscript{b}Department of Computer Science, Stanford University, Stanford, CA 94305 \\
  \textsuperscript{c}Artificial Societies \\
  \textsuperscript{d}Human-Computer Interaction Institute, Carnegie Mellon University, Pittsburgh, PA 15213
}

\date{}
\renewcommand{\undertitle}{}
\renewcommand{\headeright}{}

\hypersetup{
pdftitle={AI writing on social media is associated with greater engagement and productivity},
pdfauthor={Changyang Wu, Lynnette Hui Xian Ng, Meryl Ye, Yutong Zhang, James K. He, Felix Wallis, Steve Rathje},
pdfkeywords={generative AI, social media, engagement, LinkedIn, computational social science},
}

\begin{document}
\maketitle

\begin{abstract}
While past research has explored what goes ``viral'' or receives engagement on social media, most of this work explored virality prior to the rise of generative AI, and has not accounted for how AI has changed the nature of online engagement. Here, we explore how AI writing changes the nature of online virality using a large-scale dataset of posts on LinkedIn, a professional social media platform. Specifically, we scored 995,629 public posts by 420,288 authors using an open-weight model that estimates the extent of AI writing or editing in text. We found that an author's AI-generated or AI-edited posts received more engagement (comments and reactions) than that same author's non-AI posts. AI-assisted posts were also longer, less likely to contain links, and included more explicit requests to comment, share, and follow than non-AI posts. A mediation analysis found that these features helped explain the association between AI writing and engagement. We also followed 44,936 authors over the course of their posting histories, finding that authors who increased their AI usage (as opposed to those who did not) gained engagement per post, posted more frequently, and began writing longer posts. Despite research suggesting that people are dissatisfied with AI ``slop'' on social media, our work suggests that authors have an incentive to adopt AI in their posting in order to increase their overall engagement and productivity.
\end{abstract}


A large body of work has examined the features that drive engagement (or ``virality'') on social media \cite{rathje2025virality}. These features include emotional language \cite{berger2012viral,schone2021negativity,ng2024automated}, moral language \cite{brady2017emotion,brady2021outrage}, or identity language \cite{hopkins2025identity,rathje2021outgroup}, among others. However,  prior work on this topic has explored virality before the rise of generative AI. Recent estimates suggest that AI-generated posts are now extremely common on social media \cite{bick2024adoption,liang2025widespread,sun2025aigt}, with one 2026 estimate suggesting that more than 40\% of long-form posts on LinkedIn are AI-generated \cite{spero2026feed}.  The current paper examines how AI is reshaping the nature of social media virality.

AI detectors now make it possible to study the amount of AI in text unobtrusively. Recent work suggests that AI-generated text is statistically distinguishable from human-generated text \cite{pagan2026distinguishable}, which makes it possible to build machine learning-based models that detect statistical differences between AI-assisted and non-AI-assisted texts. Independent evaluations suggest that certain AI detectors such as Pangram are able to detect AI-generated or edited texts with few false positives \cite{jabarian2025detection}. For the current work, we used a detector called EditLens, an open-weight AI-detection model developed by Pangram and other researchers \cite{thai2026editlens}.  Several studies have applied AI detectors to large bodies of texts to trace the spread of AI-generated or assisted writing in consumer complaints, corporate press releases and job postings \cite{liang2025widespread}, scientific manuscripts \cite{kusumegi2025scientific,siler2026diffusion}, self-published books \cite{chakrabarty2026floods}, and on social media platforms \cite{sun2025aigt}. Following this work, we used EditLens to measure the use of AI in public LinkedIn posts.

It is unclear whether AI-assisted writing would be rewarded or penalized on social media. On the one hand, it is easy to see how it might be penalized. Surveys suggest that people are concerned about the proliferation of low-quality AI ``slop'' online \cite{kommers2026slop,so2026slop}, and believe that AI is making online content worse \cite{gartner2026content}. Furthermore, experiments have found that people evaluate messages and their senders more negatively when they believe AI helped write them \cite{jakesch2019aimc,hohenstein2023ai,reif2025penalty}. On TikTok, labels disclosing AI-generated content reduced engagement with posts \cite{carney2026madewithai}. People also rate news headlines labeled as AI-generated as less accurate \cite{longoni2022news,altay2024headlines} and are less willing to share them, even when the headlines are true or were written by humans \cite{altay2024headlines}. Similarly, people prefer artwork they believe was made by humans over the same artwork labeled as made by AI \cite{bellaiche2023humans}. Furthermore, since AI-generated text contains less diversity \cite{sourati2026shrinking}, people may find AI-generated social media posts less novel and interesting, and thus engage with them less. 

On the other hand, there is reason to believe that AI writing would be rewarded on social media. For instance, readers are bad at telling AI-generated from human-written text \cite{jakesch2023heuristics,paik2026humanauthored}, meaning that AI-generated text might go ``viral'' if readers cannot tell it is AI. Additionally, because AI models are trained on social media data \cite{meta2025training}, they may pick up on patterns that are associated with virality from this training data. Moreover, certain aspects of post-training, such as reinforcement learning from human feedback \cite{ouyang2022training}, which trains AI models to align with human preferences, might make AI excellent at generating clear and appealing posts that attract more engagement. Large language models are currently able to write texts that are more persuasive than humans \cite{hackenburg2026outpersuades}, that are perceived as more empathic than humans \cite{yin2024heard}, and that readers rate above human-written texts when they do not know who wrote them \cite{chakrabarty2025deepfakes,porter2024poetry,zhang2023favoritism}, so it would make sense if they also possess the ability to write more engaging social media content. AI might also indirectly increase engagement because it is able to increase an author's productivity \cite{moller2026impact}, thus giving them more opportunities to go ``viral.'' 

Here, we tested these competing hypotheses: does social media reward or penalize AI-assisted posts from a particular author? Additionally, we followed authors over time to explore how an author's engagement and productivity (how often they post and how much they write) change as they adopt AI in their posting. In addition to these within-author analyses, we looked at between-author differences in authors who use the most and least AI.

\section*{Overview}
We scored 995,629 public LinkedIn posts of at least 50 words, written by 420,288 authors between June 2025 and February 2026, with EditLens \cite{thai2026editlens}, an open-weight model that estimates how heavily a text has been edited by AI. We kept only posts of at least 50 words, because shorter texts give the detector too little text to score. We treated its output, the AI score, as a continuous measure of how heavily a post was AI-generated or AI-assisted. To help validate EditLens, we compared this recent sample of LinkedIn posts to a separate sample of posts from 2012 to 2022, finding that the EditLens score in the recent sample was substantially higher (Cohen's $d=0.91$, $P<0.001$). While EditLens has already been validated and shown to tell human-written, AI-edited, and AI-generated text apart with 90\% accuracy \cite{thai2026editlens}, we also further validated EditLens by correlating it with a dictionary-based measure of words often used by AI (Appendix, Table~\ref{tab:markers}). 

Using this dataset scored by EditLens, we first (1) looked between authors, comparing authors who used the most AI in their writing to those who used the least on various traits (e.g., topics discussed, posting country). Then, (2) we looked within authors, using multilevel models to examine whether a given author's post received more or less engagement when it had more AI-generated content, and whether such a post more often asked readers to comment, share, or follow. We also looked at the topics of AI-assisted posts, using a large language model to sort posts into 17 pre-defined topics (Methods). Furthermore, we examined the length, style, and sentiment of posts, and examined the tendency of posts to request engagement (a tactic called ``engagement bait'' \cite{facebook2017engagementbait}). Finally, we (3) looked at authors over time, to examine what happens to an author's engagement and posting history when their AI usage rises or falls. 

\section*{Results}
We measured engagement as the number of reactions (such as likes) and the number of comments a post received, and as the sum of the two. The median post received 66 reactions and 5 comments, but the most-engaged 1\% of posts received at least 5,074 reactions and comments in total (Appendix, Table~\ref{tab:descriptives}). We therefore analyzed each count as $\log(1+\text{count})$. We report results as the percentage change in engagement for every 0.1 increase in the AI score, which runs from 0 to 1. All measures and models are described in Methods.

\subsection*{Characteristics of Heavy vs. Light AI Users on Social Media}

\begin{figure}[t]
\centering
\includegraphics[width=0.49\linewidth]{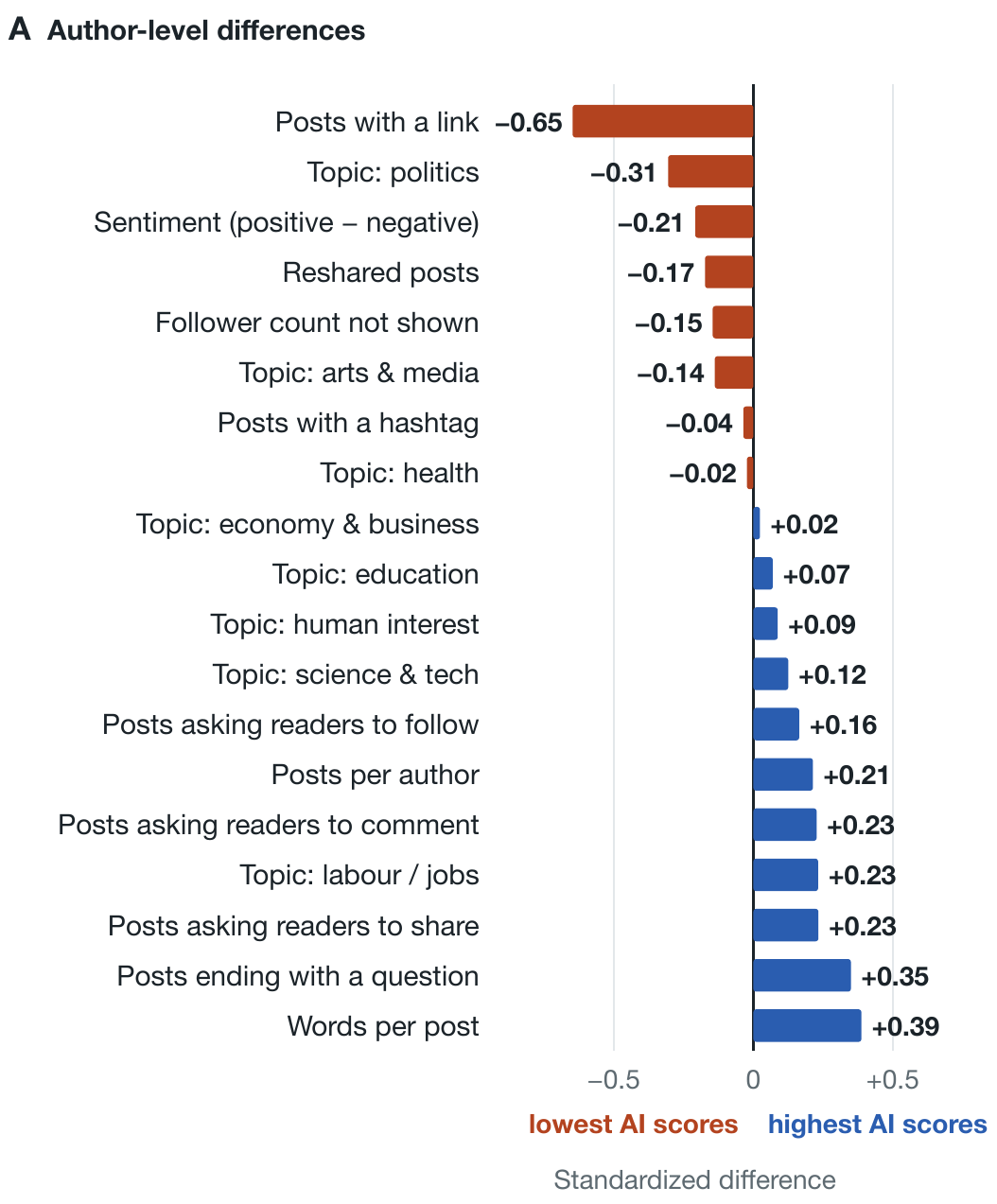}\hfill
\includegraphics[width=0.49\linewidth]{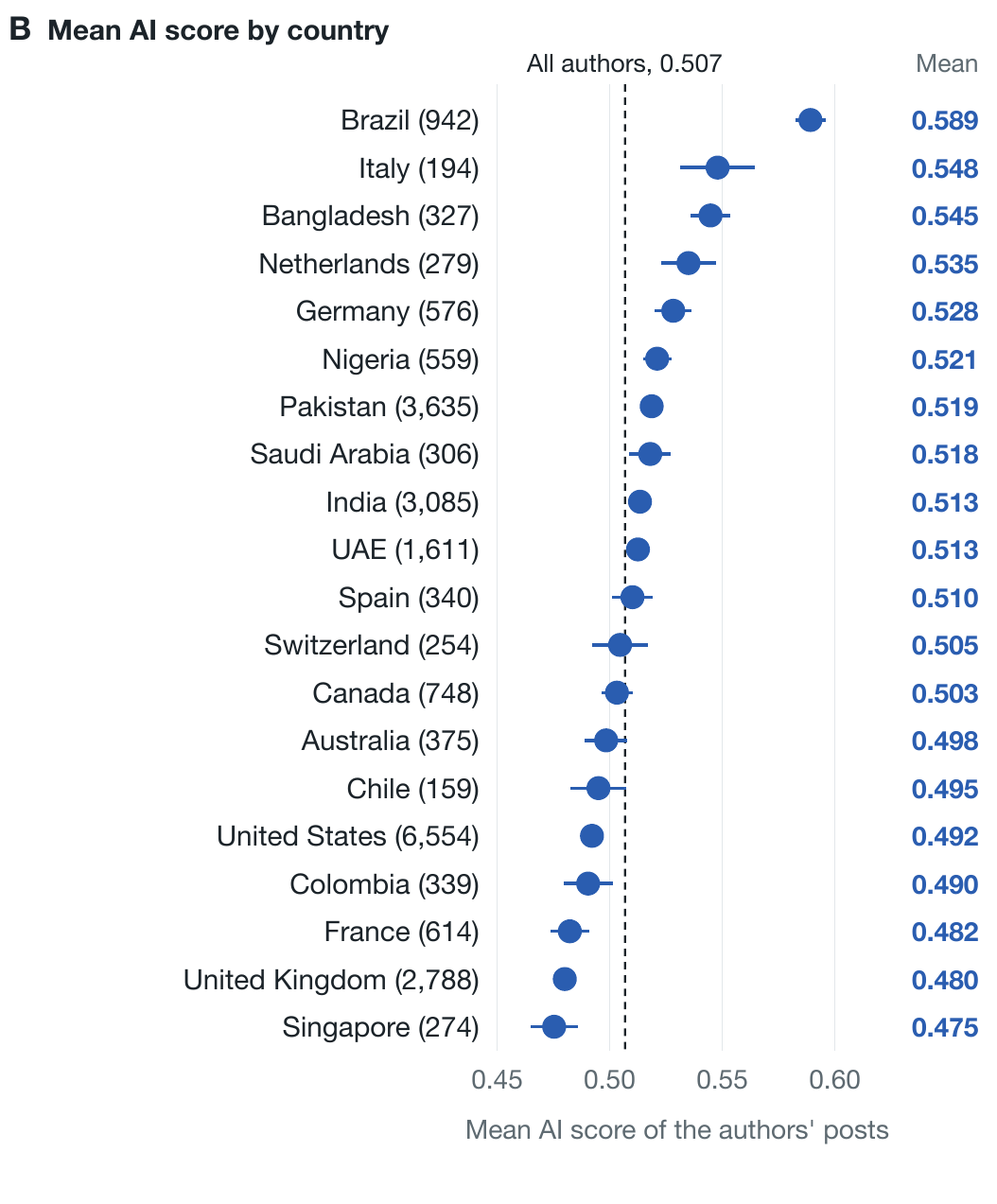}
\caption{\textbf{Who writes AI-generated posts.} The results suggest that the authors who used the most AI wrote longer posts, posted more often, asked readers for engagement more often, and included fewer links, and that AI scores were higher in countries such as Brazil, Pakistan, and India than in the US and the UK. (\textit{A}) Standardized differences between the fifths with the highest and lowest mean AI scores of authors with at least two posts (21,394 authors per fifth). Blue bars mark characteristics more common in the highest fifth, and orange bars those more common in the lowest fifth. Each characteristic is computed per post and averaged within author, except posts per author, which is a count. All differences are significant at $P<0.05$ (author-level regression with HC1 standard errors). The characteristics are defined in Appendix, Table~\ref{tab:traits_def}, and the same comparison for posts is in Appendix, Fig.~\ref{fig:post_characteristics}. (\textit{B}) Mean AI score of the same authors in each country listed on their profile, with 95\% CIs, for the 20 countries with at least 150 such authors (number of authors in parentheses). The dashed line is the mean over all 26,976 authors who listed a country.}
\label{fig:authors}
\end{figure}

To compare those who used the most and least AI in their posting, we ranked authors by the mean AI score of their posts and compared the fifths with the highest and lowest scores among authors with at least two posts (21,394 authors in each fifth; Fig.~\ref{fig:authors}\textit{A}). We left out the 313,320 authors whose average AI score would rest on a single post and therefore be less reliable. Following earlier work \cite{tan2014effect,lee2018advertising}, we detected requests to comment, share, or follow with phrases such as ``comment below'', ``repost'', and ``follow for more'' (Appendix, Table~\ref{tab:features_def}). We used a large language model to label the topic of each post with one of 17 topic categories adapted from another paper \cite{kuzman2025llm} (Appendix, Table~\ref{tab:topics_def}).

The authors who used the most AI differed substantially in the content of their posting as compared to those who used the least AI. For example, authors with the highest AI scores more often asked readers to share (5.6\% of their posts, against 2.2\%; standardized difference $d=0.23$, $t(42{,}786)=24.1$), comment (3.4\% against 1.0\%; $d=0.23$, $t(42{,}786)=23.5$), or follow (2.1\% against 0.6\%; $d=0.16$, $t(42{,}786)=17.0$; all $P<0.001$; Appendix, Fig.~\ref{fig:requests_three}\textit{B}). In other words, heavy AI users used more explicit requests for engagement in their posting. 

Heavy AI writers also wrote about different topics. For example, they wrote less about politics (1.6\% of their posts, against 5.9\%; $d=-0.31$, $t(42{,}786)=-31.6$) and more about labor and employment (18.9\% against 12.3\%; $d=0.23$, $t(42{,}786)=24.0$) and science and technology (16.6\% against 13.1\%; $d=0.12$, $t(42{,}786)=12.9$; all $P<0.001$; for posts, see Appendix, Fig.~\ref{fig:topics}). 

The heavy AI authors also posted more. Specifically, they posted 5.3 posts on average over the nine months, against 3.9 for the fifth with the lowest scores ($d=0.21$, $t(42{,}786)=22.1$, $P<0.001$). Heavy AI writers also wrote longer posts (173 against 144 words; $d=0.39$, $t(42{,}786)=40.1$; Fig.~\ref{fig:authors}\textit{A}) and were far less likely to include a link in their posts (12.3\% against 33.2\%; $d=-0.65$, $t(42{,}786)=-67.0$; both $P<0.001$).

We also examined where the heavy AI users were posting from. Among the authors in Fig.~\ref{fig:authors}\textit{A} who listed a country on their profile (26,976 of 106,968), the mean AI score was higher in Brazil (0.589), Pakistan (0.519), and India (0.513) than in the US (0.492) and the UK (0.480; Fig.~\ref{fig:authors}\textit{B} and Appendix, Table~\ref{tab:country_language}). Each of these three countries significantly differed from the US (Welch $t$ tests, $t(1{,}161)=26.6$, $t(8{,}573)=15.2$, and $t(6{,}824)=11.4$; all $P<0.001$).

\begin{figure}[t]
\centering
\includegraphics[width=0.6\linewidth]{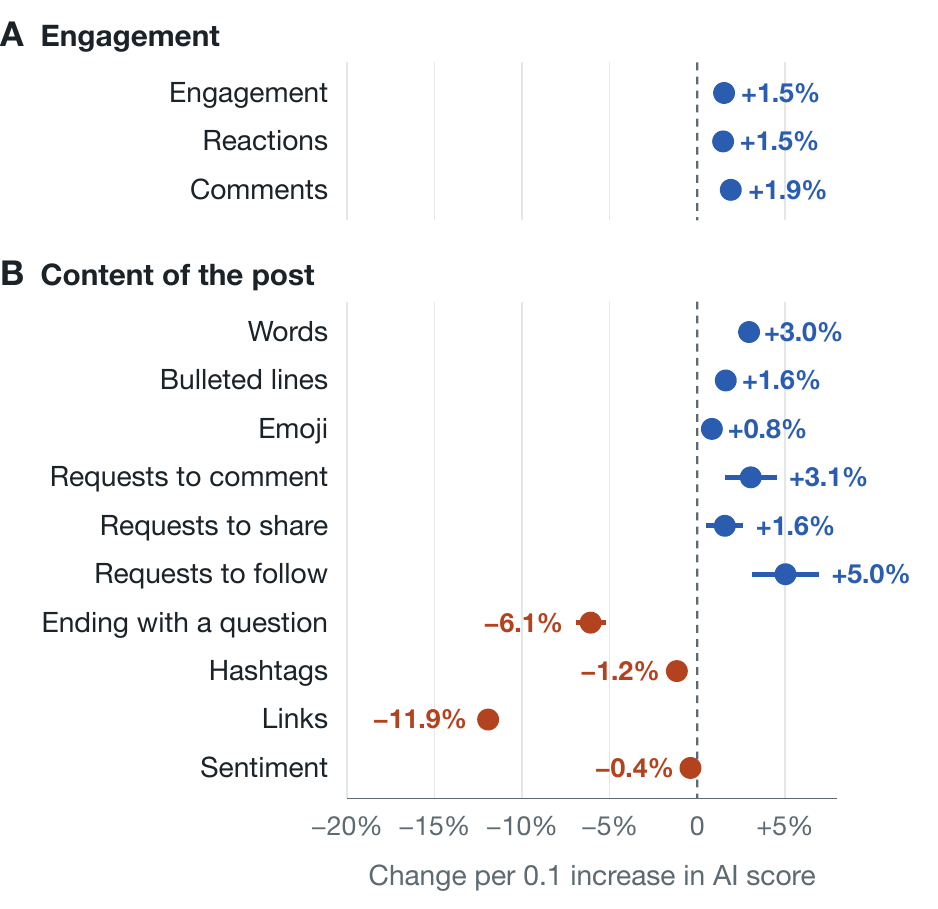}
\caption{\textbf{Engagement and content when a post had a higher AI score than the author's other posts.} Points show the percentage change in each measure for a 0.1 increase in a post's AI score, compared with the same author's other posts. Error bars represent 95\% confidence intervals. (\textit{A}) Engagement, reactions, and comments. (\textit{B}) Features of the post. The results suggest that AI-assisted posts received more engagement, reactions, and comments. They were also longer, included fewer links and more bulleted lines and emoji, and included more requests for engagement (e.g., comments, shares, and follows). All $P<0.001$ except requests to share ($P=0.004$). Estimates are from a multilevel model that holds month and weekday constant (Methods). Data are from 559,149 posts by 106,968 authors. The binned within-author relationship is shown in Appendix, Fig.~\ref{fig:engagement_bins}, and comparisons across all posts in Appendix, Fig.~\ref{fig:engagement_raw}.}
\label{fig:engagement}
\end{figure}

\subsection*{AI Writing Predicts Engagement}

\subsubsection*{Comments and Reactions} To examine whether an author's AI-assisted posts received more engagement than their non-AI posts, we used a multilevel model to compare posts by the same author, holding month and weekday constant. We found that AI writing was associated with more engagement: specifically, a 0.1-unit increase in a post's AI score (on a 0--1 scale) was associated with a 1.5\% increase in engagement ($P<0.001$). It was also associated with a 1.5\% increase in reactions, and a 1.9\% increase in comments (both $P<0.001$; Fig.~\ref{fig:engagement}). A 0.1-unit AI score increase was also associated with a 3.0\% increase in length ($P<0.001$).  Overall, this suggests that a given author's AI-assisted posts tend to be longer and receive more engagement than their non-AI posts. 

\subsubsection*{Content of the Post} We next examined how the content of an author's AI-assisted posts differed from non-AI-assisted posts (Fig.~\ref{fig:engagement}\textit{B}). Besides being longer, a post that scored 0.1 higher had 1.6\% more bulleted lines and 0.8\% more emoji (both $P<0.001$). AI-assisted posts were also more likely to ask readers to comment (3.1\% more likely for each 0.1 increase in the AI score), share (1.6\%), or follow (5.0\%; all $P\leq0.004$), and less likely to end with a question ($-6.1\%$), to contain a hashtag ($-1.2\%$), or to contain a link ($-11.9\%$; all $P<0.001$). Their sentiment was also less positive (0.4\% of a standard deviation; $P<0.001$).

\begin{table}[t]
\centering
\small
\setlength{\tabcolsep}{3pt}
\caption{\textbf{The features of a post and the response they receive.} Coefficients from a multilevel model that compares each post with the same author's other posts. The results suggest that certain features that are more likely to be in AI-assisted posts (namely, longer word count and specific requests for engagement) are also more likely to receive engagement, whereas posts with links, which are less common among AI-assisted posts, receive less engagement. The ten features of Fig.~\ref{fig:engagement}\textit{B} were entered together with the AI score, the author's follower count, and month and weekday fixed effects. The comparison across all posts is in Appendix, Table~\ref{tab:tactics_full}. Outcomes are $\log(1+\text{count})$. Words, bulleted lines, and emoji are logged counts, $\log(\text{words})$ and $\log(1+n)$; for bulleted lines and emoji, the share is of posts with at least one. Sentiment is $P(\text{positive})-P(\text{negative})$, from $-1$ to $+1$. The share of posts is out of the 559,149 posts in the within-author model. Standard errors, clustered by author, are in parentheses. Data are from 559,149 posts by 106,968 authors. $^{*}P<0.05$, $^{**}P<0.01$, $^{***}P<0.001$.}
\label{tab:tactics}
\begin{tabular}{lrrrr}
\toprule
Feature & \% of posts & Engagement & Reactions & Comments \\
\midrule
Words & --- & $+0.31^{***}$ & $+0.30^{***}$ & $+0.29^{***}$ \\
 & & (0.004) & (0.004) & (0.005) \\[2pt]
Bulleted lines & 33.2 & $-0.04^{***}$ & $-0.04^{***}$ & $-0.03^{***}$ \\
 & & (0.002) & (0.002) & (0.002) \\[2pt]
Emoji & 48.8 & $-0.01^{*}$ & $-0.01^{**}$ & $0.00$ \\
 & & (0.003) & (0.002) & (0.003) \\[2pt]
Requests to comment & 3.1 & $+0.21^{***}$ & $+0.08^{***}$ & $+0.47^{***}$ \\
 & & (0.015) & (0.011) & (0.021) \\[2pt]
Requests to share & 5.6 & $+0.09^{***}$ & $+0.08^{***}$ & $+0.10^{***}$ \\
 & & (0.009) & (0.009) & (0.010) \\[2pt]
Requests to follow & 2.4 & $+0.09^{***}$ & $+0.05^{***}$ & $+0.12^{***}$ \\
 & & (0.016) & (0.014) & (0.018) \\[2pt]
Ending with a question & 9.7 & $-0.06^{***}$ & $-0.06^{***}$ & $-0.01^{*}$ \\
 & & (0.005) & (0.005) & (0.005) \\[2pt]
Hashtags & 48.7 & $-0.04^{***}$ & $-0.04^{***}$ & $-0.08^{***}$ \\
 & & (0.005) & (0.005) & (0.005) \\[2pt]
Links & 23.0 & $-0.19^{***}$ & $-0.18^{***}$ & $-0.18^{***}$ \\
 & & (0.005) & (0.005) & (0.005) \\[2pt]
Sentiment & --- & $+0.30^{***}$ & $+0.31^{***}$ & $+0.18^{***}$ \\
 & & (0.006) & (0.006) & (0.007) \\[2pt]
AI score (0--1) & --- & $+0.01$ & $+0.01$ & $+0.05^{***}$ \\
 & & (0.013) & (0.012) & (0.014) \\
\bottomrule
\end{tabular}
\end{table}

\subsubsection*{Responses to Common Features in AI-Assisted Posts} To better understand why AI-assisted posts received more engagement than non-AI posts, we examined whether certain features that were associated with AI-assisted posts predicted engagement. First, we examined requests for engagement, which were more common in AI-assisted posts. Using a similar multilevel model comparing posts by the same author, we found that a request to comment was associated with more comments ($b=+0.47$, $P<0.001$) and more reactions ($b=+0.08$, $P<0.001$; Table~\ref{tab:tactics}). Additionally, requests to share (5.6\% of posts) and to follow (2.4\%) were associated with more of both (comments $+0.10$ and $+0.12$, reactions $+0.08$ and $+0.05$; all $P<0.001$). Ending a post with a question, which was less common in AI-assisted posts, was associated with fewer reactions ($-0.06$, $P<0.001$) and slightly fewer comments ($-0.01$, $P=0.03$). 

Longer posts also received more  engagement. Specifically, a post that was twice as long as the author's other posts received 24\% more engagement ($b=+0.31$, $P<0.001$). Posts with a link received 17\% less engagement than the same author's posts without one ($b=-0.19$, $P<0.001$), and posts with a more positive sentiment received more engagement ($b=+0.30$, $P<0.001$). In other words, many of the features that were more likely to be present in AI-assisted posts, such as a longer word count or explicit requests for engagement, were associated with more engagement (Fig.~\ref{fig:engagement}\textit{B} and Table~\ref{tab:tactics}). By contrast, ending with a question, including a hashtag, and including a link, which were less likely to be present in AI-assisted posts ($-6.1\%$, $-1.2\%$, and $-11.9\%$ per 0.1 increase in the AI score), were associated with less engagement. Bulleted lines, emoji, and sentiment were the exceptions: bulleted lines and emoji were more common in AI-assisted posts but did not go with more engagement, and AI-assisted posts were less positive, whereas more positive posts received more engagement. 

To test whether these specific features explained why AI-assisted posts received more engagement, we explored whether these features mediated the association between AI writing and engagement (Methods). Together, the ten features accounted for 92.4\% of the within-author association between the AI score and engagement, and for 72.4\% of the association with comments. Most of the association with engagement was explained by post length (59.9\%), followed by links (34.0\%) and the four requests for engagement (4.8\%), while the three formatting features ($-3.2\%$) and sentiment ($-3.1\%$) worked in the opposite direction (Appendix, Table~\ref{tab:mediation_within}). This suggests that post length and a few specific features (namely, fewer links and more requests for engagement) help explain the association between AI-assisted writing and engagement. 

\subsection*{How Engagement and Posting Behavior Change as Authors Adopt AI}

We next examined what happened as authors increased or decreased their AI usage in their posting behavior. To do this, we split each author's posting window at its midpoint and compared the first and second halves. Engagement differed from week to week, potentially because older posts had more time to collect reactions and comments. We therefore subtracted from each post's engagement, reactions, and comments how much higher or lower engagement was in that calendar week than in an average week (Methods). Because we wanted to examine authors who posted multiple times, the final sample included authors with four or more posts and one or more on each side of their midpoint: 44,936 authors, median 123 days and seven posts.

\begin{figure}[t]
\centering
\includegraphics[width=\textwidth]{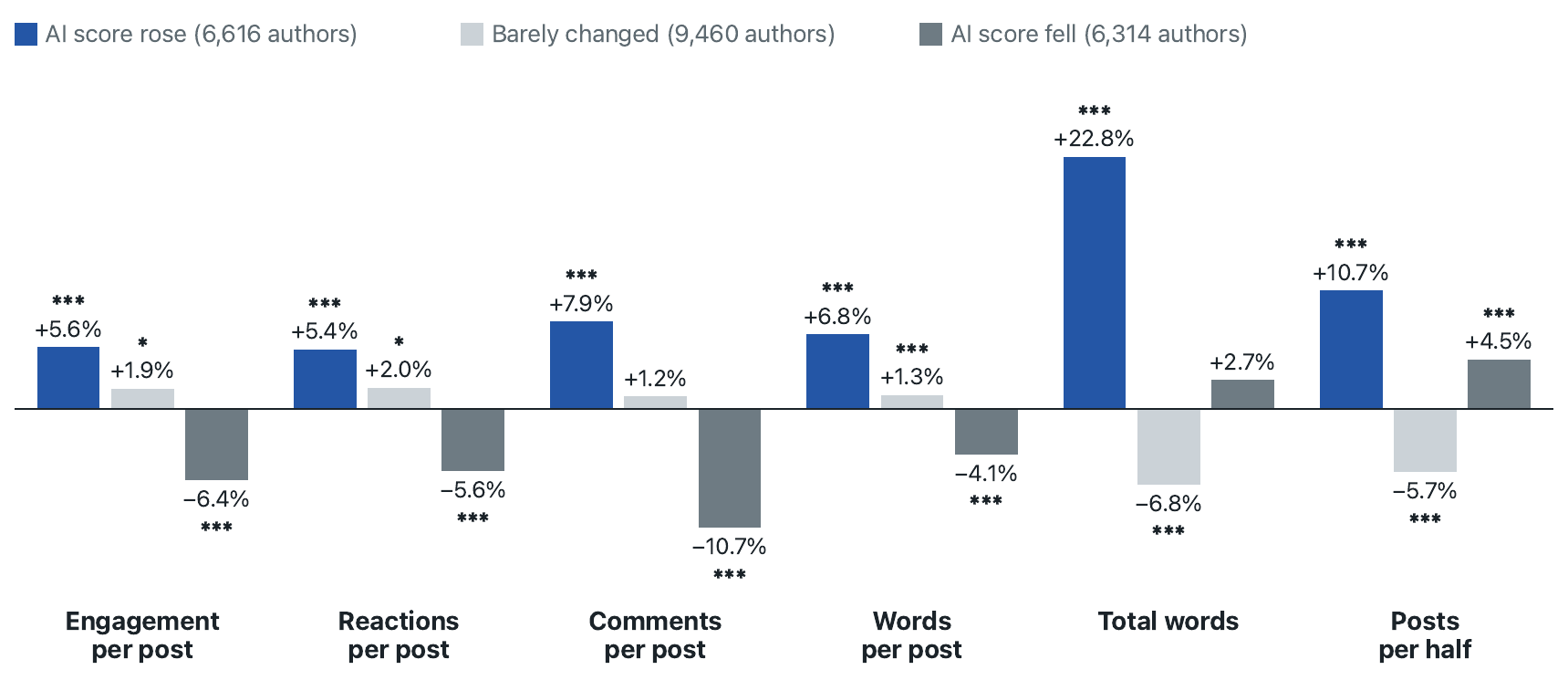}
\caption{\textbf{What happened after an author increased or decreased their AI usage.} Authors with at least four posts were split at the midpoint of their own posting window and grouped by the change in their mean AI score, which rose by at least 0.1 (blue), changed by less than 0.02 (light gray), or fell by at least 0.1 (dark gray). Bars show each group's average of its authors' own changes from the first to the second half, the change in $\log(1+\text{value})$ shown as a percentage. Engagement, reactions, and comments are net of calendar-week effects. Stars mark a change that differs significantly from zero (one-sample $t$ tests; $^{*}P<0.05$, $^{**}P<0.01$, $^{***}P<0.001$). We found that authors who increased their AI usage received more engagement per post, wrote longer posts, and posted more often. Authors who decreased their AI usage received less engagement and wrote shorter posts. Exact values are in Appendix, Table~\ref{tab:adoption_groups}. Estimates per 0.1 change in the author's mean AI score, alone and with the changes in post length, sentiment, and number of posts held constant, are in Appendix, Table~\ref{tab:adoption}.}
\label{fig:adoption}
\end{figure}

We grouped authors by the change in their mean AI score between the first and second half. Specifically, we compared authors who increased their mean AI score by 0.1 points or more (on a 0--1 scale; 6,616 authors, from 0.412 to 0.579 on average) with authors who decreased it by 0.1 points or more (6,314 authors, from 0.580 to 0.414) and with authors whose score changed by less than 0.02 points (9,460 authors). For each group, we report how engagement and posting changed from the first half to the second (Fig.~\ref{fig:adoption}).

\subsubsection*{Changes in Engagement and Posting} We first compared how engagement changed among those who increased, decreased, or stayed the same in their AI usage (Fig.~\ref{fig:adoption}). Authors whose AI scores rose gained engagement: in the second half, their posts received 5.6\% more engagement than in the first half ($t(6{,}615)=4.3$, $P<0.001$) and 7.9\% more comments ($t(6{,}615)=5.1$, $P<0.001$). By contrast, those whose AI scores fell lost engagement, receiving 6.4\% less engagement ($t(6{,}313)=-5.1$, $P<0.001$), 5.6\% fewer reactions ($t(6{,}313)=-4.5$, $P<0.001$), and 10.7\% fewer comments ($t(6{,}313)=-7.3$, $P<0.001$). Authors whose AI scores barely changed saw little change in engagement ($+1.9\%$, $t(9{,}459)=2.4$, $P=0.019$) or comments ($+1.2\%$, $t(9{,}459)=1.3$, $P=0.18$; Appendix, Table~\ref{tab:adoption_groups}).

We next examined whether authors wrote and posted more as their AI scores rose. Authors whose AI scores rose wrote 6.8\% more words per post in the second half ($t(6{,}615)=11.1$, $P<0.001$), whereas authors whose scores fell wrote 4.1\% fewer ($t(6{,}313)=-6.9$, $P<0.001$) and authors whose scores barely changed wrote 1.3\% more ($t(9{,}459)=3.6$, $P<0.001$). Because the two halves of an author's posting window are equally long, the number of posts in a half is a posting rate. Authors whose AI scores rose posted 10.7\% more often in the second half ($t(6{,}615)=10.8$, $P<0.001$), against 4.5\% more for authors whose scores fell ($t(6{,}313)=4.6$, $P<0.001$) and 5.7\% less for authors whose scores barely changed ($t(9{,}459)=-7.6$, $P<0.001$). In total, authors whose AI scores rose wrote 22.8\% more words in the second half ($t(6{,}615)=13.7$, $P<0.001$), against 2.7\% more for authors whose scores fell ($t(6{,}313)=1.7$, $P=0.087$) and 6.8\% fewer for authors whose scores barely changed ($t(9{,}459)=-6.5$, $P<0.001$; Appendix, Table~\ref{tab:adoption_groups}). These results were robust to looser and stricter sample definitions (Appendix, Fig.~\ref{fig:S7} and Table~\ref{tab:adoption_samples}), and the differences between the groups are in Appendix, Table~\ref{tab:adoption_groups}.

\subsubsection*{Continuous Change in the AI Score} Because the prior analysis focused on comparisons between groups, we also examined the continuous association between an author's change in mean AI score and their engagement and posting. Across all 44,936 authors, a 0.1 rise in an author's mean AI score between halves was associated with 3.8\% more engagement per post ($t(44{,}934)=8.0$), 3.6\% more reactions ($t(44{,}934)=7.6$), and 6.1\% more comments ($t(44{,}934)=10.2$; all $P<0.001$; Appendix, Table~\ref{tab:adoption}). It was also associated with 3.1\% more words per post ($t(44{,}934)=13.9$), 1.7\% more posts ($t(44{,}934)=4.8$), and 5.3\% more words written in total ($t(44{,}934)=9.4$; all $P<0.001$). When we held constant the changes in post length, sentiment, and number of posts, the association was smaller but remained: 2.4\% for engagement ($t(44{,}931)=5.1$) and 4.4\% for comments ($t(44{,}931)=7.6$; both $P<0.001$).

\section*{Discussion}
Using a large dataset of nearly one million LinkedIn posts, we examined whether audiences reward or penalize AI-generated or AI-assisted writing on social media. We found that AI-assisted posts from a given author receive more engagement than non-AI-assisted posts from that same author. We also found that AI-assisted posts were longer than non-AI-assisted posts, contained fewer links, and contained more explicit requests for engagement (e.g., requests for comments, shares, and follows). Longer posts and posts with these requests received more engagement, whereas posts with links received less, and a mediation model found that together these features explained most of the link between AI usage and engagement. This suggests that AI-assisted posts from a given author go more ``viral'' than non-AI-assisted posts from that same author, in part because they contain more features that are associated with virality. 

When we followed authors over time, we found that authors who increased their AI usage (as compared to those who decreased their AI usage or did not change) received more engagement per post. They also posted more and began writing longer posts. This suggests that social media content creators now have an incentive to use AI in their posting in order to increase their engagement and productivity. This mirrors other work suggesting that AI makes people more productive in other domains, such as science \cite{kusumegi2025scientific}.

We also looked at between-author differences between heavy and light AI users on social media. We found that heavy AI users were more likely to reside outside of the US and the UK, and were more likely to write about specific topics (labor and employment, and science and technology, as opposed to politics), were less likely to include links in their posts, wrote longer posts, and included more requests for engagement in their posts. 

There are multiple reasons why AI writing might increase engagement. For instance, it might boost engagement by increasing productivity. Indeed, supporting this explanation, we found those who increased their AI usage became more productive over time. Moreover, the strongest mediator between AI usage and engagement was post length, suggesting that AI-assisted posts may go more viral simply because authors are able to write more. On the other hand, AI-assisted writing may also contain characteristics that are associated with engagement. Supporting this explanation, we also found that more requests for engagement explained part of the association between AI writing and engagement. It should be noted that our work is correlational, so it is possible that a third variable may explain these results (e.g., authors who care more about a post going viral are more likely to use AI to improve it). Future work should follow up on these results with causal investigations. 

While AI-assisted posts from an author receive more engagement than non-AI-assisted posts from that same author, that does not necessarily mean those posts are higher quality. Indeed, prior work suggests that widely shared content is often not widely liked \cite{rathje2024divisive}. Moreover, AI-assisted posts more often include explicit requests for engagement \cite{zhang2023cue}, which are tactics that many platforms explicitly demote in recommendation algorithms as they are considered spammy \cite{facebook2017engagementbait,watti2023engagementbait,linkedin2026avoid}. Moreover, public opinion polls suggest that people are concerned about AI ``slop'' \cite{gartner2026content}, and LinkedIn has even recently implemented a feature that allows users to report AI slop \cite{duffy2026linkedin}. Thus, it is possible that AI is making individual content creators more productive and successful while degrading the overall social media experience. 

Our work has limitations that should be addressed in future work. For instance, we could only predict AI-assisted writing using a classifier, not directly observe it. However, the classifier we used has been shown to be highly accurate \cite{thai2026editlens}, and we independently validated it against a sample of pre-ChatGPT LinkedIn data and a measure of words often used by AI. Moreover, our data are observational, so we cannot make causal claims about the impact of AI writing on engagement. Further, our work focuses on one platform, LinkedIn, in part because it is one of the few platforms that includes long posts, which are required for accurate AI classification. Future work should focus on replicating these patterns across platforms. Finally, because our work focuses on undisclosed AI use, future work should examine whether tools that disclose AI usage (such as AI labels, or browser extensions that reveal AI text) change the engagement patterns we see here. 

\section*{Conclusions}

Despite research suggesting that people are dissatisfied with ``AI slop'' on social media, our work suggests that undisclosed AI usage is rewarded with more engagement. Because AI use is associated with increased productivity and engagement, authors may have an incentive to adopt it in their social media posting—even if doing so degrades the overall social media experience.

\section*{Materials and Methods}
\subsection*{LinkedIn Posts}
We collected 1,401,505 public LinkedIn posts by 579,253 authors, published between 1 June 2025 and 26 February 2026. The posts come from a database of public LinkedIn activity kept by Reverse Contact, a commercial data provider, from which we drew a random sample of up to 200,000 posts per month. Author identifiers were hashed first, to keep authors anonymous. For each author we then retrieved the posts that author had recently published, commented on, or reacted to. We kept posts of at least 50 words, counted by whitespace (995,629 posts by 420,288 authors), so that each text was long enough to score. The regressions leave out 116,934 repeated scrapes of the same post (same author and text) and 6,226 posts without an author identifier, which leaves 872,469 posts. The within-author comparisons use the 559,149 of these posts that were written by the 106,968 authors with at least two posts. Over all 995,629 posts, the mean engagement was 373 reactions and comments and the median 76; 29.3\% of authors posted two or more times (median one post). Sample construction and descriptive statistics are given in Appendix, Tables~\ref{tab:data_collected} and \ref{tab:descriptives}. To validate the AI score, we also scored 204,468 LinkedIn posts published between October 2012 and March 2022, before the public release of ChatGPT, of which 109,355 had at least 50 words.

\subsection*{AI Scores}
We used EditLens \cite{thai2026editlens}, an open-weight model that estimates how heavily a text has been edited by AI, to score each post. We used the base Llama-3.2-3B model with the \texttt{pangram/\allowbreak editlens\_\allowbreak Llama-3.2-3B} score head, loaded in 4-bit NF4 quantization, and scored up to 1,024 tokens of each post. EditLens assigns each text a probability for each of four ordered classes, ranging from human-written (class 0) to AI-generated (class 3). We then multiplied each class index (0 to 3) by its probability, summed, and divided the sum by three to get a score between 0 and 1. This is the expected class divided by the highest class, the decoding proposed by the EditLens authors and used in their released inference code \cite{editlens2025code}. In our data, the mean score was 0.504 ($SD=0.131$).

\subsection*{Engagement Measures}
We analyzed reactions and comments separately and summed, each as $\log(1+\text{count})$. The data contained no share counts and no breakdown of reactions by type.

\subsection*{Sentiment Analysis}
We estimated the sentiment valence of each post, $P(\text{positive})-P(\text{negative})$, with a multilingual XLM-RoBERTa model trained on tweets (\texttt{cardiffnlp/\allowbreak twitter-xlm-roberta-base-sentiment}). Long posts were split into chunks, and the chunk probabilities were weighted by length.

\subsection*{Topic Classification}
We assigned each post a primary topic from a closed set of 17 categories with \texttt{gpt-5.6-luna}, through the OpenAI Batch API with structured outputs (categories and definitions in Appendix, Table~\ref{tab:topics_def}). The categories are the top level of the IPTC Media Topics taxonomy \cite{iptc2026mediatopics}, and we took the list from \texttt{classla/\allowbreak multilingual-\allowbreak IPTC-\allowbreak news-\allowbreak topic-\allowbreak classifier}, a multilingual news topic classifier that uses them \cite{kuzman2025llm}. Every post was labeled, and \texttt{gpt-5.6-luna} rated 89.4\% of its labels as high confidence.

\subsection*{Controls}
The main post-level models held constant when a post was published (calendar-month and weekday fixed effects) and the size of the author's audience, $\log(1+\text{followers})$ with an indicator for a missing follower count (missing for 63.2\% of posts). We did not hold constant characteristics of the post itself, such as its length and formatting, because these change with the AI score (Fig.~\ref{fig:engagement}\textit{B}).

\subsection*{Requests for Engagement and Formatting}
Following earlier work that identified explicit requests for engagement from specific phrases in a post \cite{tan2014effect,lee2018advertising}, we detected seven features with regular expressions (patterns in Appendix, Table~\ref{tab:features_def}, including German, Dutch, Spanish, French, and Portuguese forms). Four were solicitations: an explicit request to comment, to share or repost, or to follow, and a question at the end of the post, that is, a question mark in the post's closing line, ignoring links and lines of only hashtags or emoji (for a post written as one paragraph, in its last sentence). Examples of each from the data are given in Table~\ref{tab:request_examples}. Three were formatting features: any hashtag, $\log(1+\text{emoji})$ and $\log(1+\text{bulleted lines})$.

\begin{table}[t]
\centering
\small
\captionsetup{singlelinecheck=false}
\caption{\textbf{Examples of requests for engagement.}}
\label{tab:request_examples}
\begin{tabular}{@{}>{\raggedright\arraybackslash}p{0.19\linewidth}>{\centering\arraybackslash}p{0.62\linewidth}>{\centering\arraybackslash}p{0.1\linewidth}@{}}
\toprule
Request & Example & Share of posts \\
\midrule
Request to comment & This is the only thing you need to make money on LinkedIn. And I promise: it'll blow your mind. \textbf{Comment ``TRUST'' and I'll send it to you} & 3.1\% \\[4pt]
Request to share or repost & Every small step today, will lead to a big impact tomorrow. And I am here to support YOU. \textbf{Repost to help your network.} & 5.6\% \\[4pt]
Request to follow or connect & Imagine if every parking lot in your city became a mini solar power station. Would you support that? \textbf{Follow me for more content on sustainability.} & 2.4\% \\[4pt]
Question at the end & My advice for anyone hiring: 1.~Select the right people. 2.~Agree on deliverables (be crystal clear). 3.~Provide proper tools and support. 4.~Get out of their way. \textbf{Agree?} & 9.7\% \\
\bottomrule
\end{tabular}

{\vspace{4pt}\raggedright\sffamily\small\hspace*{9pt}One excerpt for each request in Appendix, Table~\ref{tab:features_def}, chosen from the 60 English-language posts with the most reactions and comments that contained the request and no other request to comment, share, or follow. Excerpts are shortened, with line breaks, emoji, and divider lines removed, and the request is in bold. The share of posts is the share of the 559,149 posts in the within-author model that contained the request.\par}
\end{table}

\subsection*{Author Comparisons}
We averaged the AI scores of each author's posts. Among authors with at least two posts, we compared the quintiles with the highest and lowest mean AI scores. For each trait, we report the standardized difference between these quintiles, the difference in means divided by the pooled standard deviation. We tested it with an author-level OLS regression on a top-quintile indicator (HC1 standard errors). For posts per author, the trait is the author's number of posts. Among the 106,968 authors with at least two posts, we compared quartiles of the mean AI score by the country listed on the profile, and compared each country with the US using Welch $t$ tests.

\subsection*{Location and Language}
We used the location field of each author's LinkedIn profile, which in our data includes a country when the author listed one. For Appendix, Table~\ref{tab:country_language}, we detected the language of every post by authors with at least two posts who listed a country, using the Python package \texttt{langdetect} on the first 600 characters of the post (270,315 posts, 89.7\% in English), and averaged each author's AI score over all of their posts and over their English posts only. For each country, we calculated the share of its authors in each quartile of the mean AI score among all 106,968 authors with at least two posts.

\subsection*{Post Comparisons}
We compared engagement across quartiles of the post's AI score (cut points 0.414, 0.502, and 0.594) and across deciles. We compared 14 post characteristics between the fifths of posts with the highest and lowest AI scores ($N=199{,}126$ each), using standardized differences and Welch $t$ tests (Appendix, Fig.~\ref{fig:post_characteristics}). We compared the distribution of primary topics between these two fifths with a $\chi^2$ test and Cramér's $V$ (Appendix, Fig.~\ref{fig:topics}).

\subsection*{Engagement Regressions}
We estimated
\[
\log(1+y_{it}) = \beta\,\text{AI}_{it} + \gamma'\mathbf{x}_{it}
  + \alpha_i + \tau_{m(t)} + \delta_{d(t)} + \varepsilon_{it}
\]
for post $t$ by author $i$: a regression of $\log(1+y)$ on the AI score, with calendar-month ($\tau$) and day-of-week ($\delta$) fixed effects. In the across-post model, $\mathbf{x}_{it}$ holds the follower controls and there is no author effect, so posts are compared across authors. In the within-author model, $\alpha_i$ is a random intercept for each author, and the AI score is split into the author's mean and the post's deviation from it; the coefficient on the deviation compares each post only with the same author's other posts, and the author mean takes up differences between authors. In the request models, each of the ten features of Fig.~\ref{fig:engagement}\textit{B} (the seven features above, log word count, whether the text contained a link, and sentiment) is split the same way. This multilevel model was fitted by restricted maximum likelihood with the follower controls and month and weekday fixed effects. The 313,320 posts by authors with only one post carry no within-author information and are not used. Author fixed-effects regressions give the same within-author estimates (to within 0.003) and are used for the robustness checks. The across-post model was fitted by ordinary least squares, with the fixed effects absorbed by alternating projections (Appendix, Table~\ref{tab:full_models}). Standard errors are clustered by author throughout. Coefficients describe a change in the AI score from 0 to 1. For Fig.~\ref{fig:engagement}\textit{B}, each of the ten features was in turn the outcome of the within-author model. Fig.~\ref{fig:engagement} shows the coefficients as percentage changes for a 0.1 increase in the AI score. For words, this is the change in the count, and for engagement, reactions, comments, bulleted lines, and emoji, the change in one plus the count. For the yes/no features, it is the change relative to how common the feature was: requests to comment were in 3.1\% of these posts, to share 5.6\%, to follow 2.4\%, ending with a question 9.7\%, hashtags 48.7\%, and links 23.0\%. For sentiment, it is in percent of a standard deviation. We also added terms in seven steps, from the AI score alone (M1) through log word count, sentiment, the platform controls (whether the post was republished, follower count, whether the text contained a URL, and $\log(1+\text{hashtags})$), month and weekday, and topic fixed effects (M6), to author fixed effects (M7; Appendix, Table~\ref{tab:ladder_spec}). With all of them added, the within-author association with engagement was close to zero and the association with comments was smaller. Post length and links accounted for most of the change. Robustness checks of the main models are in Appendix, Table~\ref{tab:robustness}.

\subsection*{Mediation Analysis}
We tested how much of the association between the AI score and engagement, and between the AI score and comments, ran through these ten features. We used a product-of-coefficients approach. We added the ten features to the across-post model and to the within-author multilevel model, and fitted three equations, for each feature on the AI score, for the outcome on the AI score and the features, and for the outcome on the AI score alone,
\begin{align*}
M_{k,it} &= a_k\,\text{AI}_{it} + \ldots,\\
y_{it} &= c'\,\text{AI}_{it} + \textstyle\sum_k b_k M_{k,it} + \ldots,\\
y_{it} &= c\,\text{AI}_{it} + \ldots
\end{align*}
with the same controls, fixed effects, and sample. As a result, $b_k$ and $c'$ are the coefficients of the model with the features, and the indirect effects $a_k b_k$ sum to $c-c'$, exactly across posts and to within 0.001 in the multilevel models. We estimated confidence intervals for the indirect effects with 1,000 bootstrap samples of authors (percentile intervals). Across posts, we partialled out the controls and fixed effects on the full sample, and within authors we refitted the multilevel models on each bootstrap sample (Appendix, Tables~\ref{tab:mediation} and \ref{tab:mediation_within}).

\subsection*{Change Within Authors Over Time}
We split each author's posting window, from their first to their last post, at its midpoint in time, and computed each quantity as the change from the first half to the second. We included authors with at least four posts whose first and last posts were not simultaneous and who had at least one post in each half ($N=44{,}936$ authors; median seven posts over 123 days). For each author and half, we computed the mean AI score, the mean $\log(1+x)$ engagement, reactions, and comments (net of the calendar week), and $\log(1+\text{words per post})$, as well as $\log(1+x)$ of the number of posts and of the total words written, and the share of posts that asked readers to comment, share, or follow. We regressed the changes in these outcomes on the change in the author's mean AI score, first alone and then with the changes in log word count, sentiment, and number of posts as controls, using OLS with HC3 standard errors. Appendix, Table~\ref{tab:adoption} shows these estimates as percentage changes for a 0.1 change in the score. We also compared authors whose mean score rose by at least 0.1 ($N=6{,}616$), authors whose score fell by at least 0.1 ($N=6{,}314$), and authors whose score changed by less than 0.02 ($N=9{,}460$). A change of 0.1 is about one standard deviation of the change across all 44,936 authors (0.105). For each group, a one-sample $t$ test compared the average change with zero (the stars in Fig.~\ref{fig:adoption}), and Welch $t$ tests compared the groups with each other (Appendix, Table~\ref{tab:adoption_groups}). Average changes in log outcomes are shown as $100\,[\exp(\text{mean change})-1]\%$. We repeated these analyses for authors with at least two and at least three posts, and for authors with at least four posts and at least two in each half (Appendix).

\subsection*{Calendar-Week Adjustment}
Before computing these changes, we removed calendar-week effects from engagement, reactions, and comments. Engagement keeps accruing after a post is published, and the platform's engagement level changed over the nine months: relative to an average week, $\log(1+\text{engagement})$ was 0.12 to 0.21 higher in the weeks of June 2025 and 0.47 lower in the week of 9 February 2026, the busiest week in the data (81,005 posts). For each of $\log(1+\text{engagement})$, $\log(1+\text{reactions})$, and $\log(1+\text{comments})$, we fitted a model with author and calendar-week (Monday to Sunday) fixed effects on all 988,150 posts with an author identifier, using alternating projections. We centered the week effects at zero over posts and subtracted each post's week effect from its value. The AI score, word counts, and numbers of posts were not adjusted. The two halves of an author's window are equally long, so the number of posts in a half is a posting rate.

\subsection*{Validation of EditLens}
To validate the AI score, we used the 109,355 LinkedIn posts of at least 50 words published between October 2012 and March 2022, before the public release of ChatGPT (described above). The mean score was 0.390 for these posts and 0.504 for the posts from 2025--26 (Cohen's $d=0.91$). In the older posts, 6.56\% fell in the model's highest class (AI-generated, a false positive for text written before ChatGPT) and 41.9\% in its lowest (human-written); in the 2025--26 posts, the figures were 23.16\% and 15.3\%. Within authors, posts with higher scores also contained more of the surface markers often seen in LLM text (all $P<0.001$; Appendix, Table~\ref{tab:markers}).

\paragraph{Acknowledgments.} This work was supported by National Science Foundation grant 2616250 (to S.R.). We thank \href{https://brightinitiative.com}{The Bright Initiative powered by Bright Data} for providing the LinkedIn posts from 2012 to 2022.

\bibliography{references}

\clearpage

\raggedbottom
\section*{Appendix}

\setcounter{figure}{0}\setcounter{table}{0}
\renewcommand{\thefigure}{S\arabic{figure}}\renewcommand{\thetable}{S\arabic{table}}
\renewcommand{\theHfigure}{S\arabic{figure}}\renewcommand{\theHtable}{S\arabic{table}}

\section*{LinkedIn sample and models}

\begin{table}[H]
    \centering
    \caption{\textbf{Data collected.} The 872,469 posts by 420,288 authors are the estimation sample for the post-level models, and the next two rows are the subset used for within-author comparisons. The over-time analysis uses all posts of at least 50 words.}
    \label{tab:data_collected}
    \begin{tabular}{lr}
    \toprule
     & \textbf{LinkedIn} \\ \midrule
     Raw posts collected & 1,401,505 \\
     \# authors of raw posts & 579,253 \\
     Posts of $\geq50$ words & 995,629 \\
     \# authors of posts $\geq50$ words & 420,288 \\
     After removing repeated posts (same author, same text) & 878,695 \\
     After removing posts with no author identifier & 872,469 \\
     \quad of which by authors with $\geq2$ posts (within-author models) & 559,149 \\
     \quad \# such authors & 106,968 \\
     Authors in the over-time analysis ($\geq4$ posts, one in each half) & 44,936 \\
     Collection window & 1 Jun 2025 -- 26 Feb 2026 \\
    \bottomrule
    \end{tabular}
\end{table}

\begin{table}[H]
    \centering
    \small
    \caption{\textbf{The 995,629 posts of at least 50 words, by 420,288 authors.}}
    \label{tab:descriptives}
    \begin{tabular}{lrrr}
    \toprule
     & Mean & Median & Highest \\ \midrule
    Posts per author & 2.35 & 1 & 765 \\
    Words per post & 159 & 133 & 466 (99th percentile) \\
    Reactions plus comments & 373 & 76 & 5,074 (99th percentile) \\
    Reactions & & 66 & \\
    Comments & & 5 & \\
    AI score & 0.504 (SD 0.131) & & \\
    Sentiment & 0.159 (SD 0.410) & & \\
    \midrule
     & \multicolumn{3}{r}{Share of posts} \\ \midrule
    No reactions or comments & \multicolumn{3}{r}{1.2\%} \\
    Republished & \multicolumn{3}{r}{11.9\%} \\
    Follower count of the author missing & \multicolumn{3}{r}{62.5\%} \\
    Author lists a location & \multicolumn{3}{r}{39.6\%} \\
    English (stop-word ratio $\geq0.12$) & \multicolumn{3}{r}{86.7\%} \\
    Topic label rated high confidence & \multicolumn{3}{r}{89.4\%} \\
    \midrule
    Authors with two or more posts & \multicolumn{3}{r}{123,182 (29.3\%)} \\
    \bottomrule
    \end{tabular}
\end{table}

\begin{table}[H]
    \centering
    \small
    \caption{\textbf{The main models and the model ladder.} Coefficients on the AI score (0 to 1), with standard errors clustered by author in parentheses. Outcomes are $\log(1+\text{count})$. The first two rows are the main models (all coefficients in Table~\ref{tab:full_models}). The ladder below adds terms one step at a time, ending with author fixed effects (M7). The last three rows are author fixed-effects regressions with the controls of the main models plus log word count, whether the text contained a URL, or both. The within-$R^2$ is 0.278 at M6 and 0.056 at M7. Robustness checks of the main models are in Table~\ref{tab:robustness}.}
    \label{tab:ladder_spec}
    \resizebox{\textwidth}{!}{%
    \begin{tabular}{llrrrr}
    \toprule
    \textbf{Model} & \textbf{Adds} & \textbf{$N$ posts / authors} & \textbf{Engagement} & \textbf{Reactions} & \textbf{Comments} \\ \midrule
    Main, across posts & follower count, month and weekday fixed effects & 872,469 / 420,288 & $+0.085$ (0.021) & $-0.057$ (0.021) & $+0.953$ (0.021) \\
    Main, within author & multilevel model, AI score split into author mean and deviation (deviation shown) & 559,149 / 106,968 & $+0.152$ (0.013) & $+0.146$ (0.013) & $+0.189$ (0.014) \\ \midrule
    M1 & AI score alone & 872,469 / 420,288 & $-0.123$ (0.030) & $-0.294$ (0.030) & $+0.970$ (0.027) \\
    M2 & + $\log$ word count & 872,469 / 420,288 & $-0.310$ (0.029) & $-0.454$ (0.029) & $+0.667$ (0.026) \\
    M3 & + sentiment & 872,469 / 420,288 & $-0.334$ (0.029) & $-0.480$ (0.029) & $+0.665$ (0.026) \\
    M4 & + platform controls & 872,469 / 420,288 & $-0.281$ (0.020) & $-0.380$ (0.020) & $+0.417$ (0.020) \\
    M5 & + month and weekday fixed effects & 872,469 / 420,288 & $-0.269$ (0.021) & $-0.366$ (0.020) & $+0.421$ (0.020) \\
    M6 & + topic fixed effect & 872,469 / 420,288 & $-0.263$ (0.020) & $-0.346$ (0.020) & $+0.337$ (0.019) \\
    M7 & + author fixed effect & 559,149 / 106,968 & $+0.005$ (0.013) & $+0.006$ (0.012) & $+0.043$ (0.014) \\ \midrule
    Within author & main controls + $\log$ word count & 559,149 / 106,968 & $+0.090$ (0.013) & $+0.088$ (0.012) & $+0.123$ (0.014) \\
    Within author & main controls + URL & 559,149 / 106,968 & $+0.097$ (0.013) & $+0.094$ (0.013) & $+0.139$ (0.014) \\
    Within author & main controls + $\log$ word count and URL & 559,149 / 106,968 & $+0.034$ (0.013) & $+0.034$ (0.012) & $+0.070$ (0.014) \\
    \bottomrule
    \end{tabular}}
\end{table}

\begin{table}[H]
\centering
\small
\caption{\textbf{All coefficients of the main engagement models.} Across posts, an ordinary least-squares regression with the AI score, the author's follower count, an indicator for a missing count, and month and weekday fixed effects. Within author, a multilevel model with a random intercept for each author and the AI score split into the author's mean and the post's deviation from it. Missing follower counts are set to zero and marked by the indicator. Outcomes are $\log(1+\text{count})$. Standard errors, clustered by author, are in parentheses. $^{*}P<0.05$, $^{**}P<0.01$, $^{***}P<0.001$.}
\label{tab:full_models}
\resizebox{\textwidth}{!}{%
\begin{tabular}{lrrrrrr}
\toprule
 & \multicolumn{3}{c}{Across posts} & \multicolumn{3}{c}{Within author} \\
\cmidrule(lr){2-4}\cmidrule(lr){5-7}
 & Engagement & Reactions & Comments & Engagement & Reactions & Comments \\
\midrule
AI score (0--1); within author: deviation from the author's mean & $+0.085^{***}$ & $-0.057^{**}$ & $+0.953^{***}$ & $+0.152^{***}$ & $+0.146^{***}$ & $+0.189^{***}$ \\
 & \footnotesize{(0.021)} & \footnotesize{(0.021)} & \footnotesize{(0.021)} & \footnotesize{(0.013)} & \footnotesize{(0.013)} & \footnotesize{(0.014)} \\[2pt]
Author's mean AI score & --- & --- & --- & $-0.137^{***}$ & $-0.361^{***}$ & $+1.265^{***}$ \\
 &  &  &  & \footnotesize{(0.041)} & \footnotesize{(0.040)} & \footnotesize{(0.039)} \\[2pt]
$\log(1+\text{followers})$ & $+0.726^{***}$ & $+0.716^{***}$ & $+0.579^{***}$ & $+0.704^{***}$ & $+0.693^{***}$ & $+0.558^{***}$ \\
 & \footnotesize{(0.005)} & \footnotesize{(0.005)} & \footnotesize{(0.006)} & \footnotesize{(0.004)} & \footnotesize{(0.004)} & \footnotesize{(0.004)} \\[2pt]
Follower count missing & $+6.959^{***}$ & $+6.963^{***}$ & $+4.923^{***}$ & $+6.487^{***}$ & $+6.465^{***}$ & $+4.582^{***}$ \\
 & \footnotesize{(0.045)} & \footnotesize{(0.045)} & \footnotesize{(0.051)} & \footnotesize{(0.036)} & \footnotesize{(0.036)} & \footnotesize{(0.033)} \\[2pt]
\midrule
$R^2$ (across) / author share of variance (within) & 0.241 & 0.251 & 0.144 & 0.692 & 0.700 & 0.611 \\
Posts / authors & \multicolumn{3}{c}{872,469 / 420,288} & \multicolumn{3}{c}{559,149 / 106,968} \\
\bottomrule
\end{tabular}}
\end{table}

\begin{table}[H]
\centering
\small
\caption{\textbf{Robustness of the AI coefficient.} Coefficients on the AI score from the across-post model and from a within-author model with author fixed effects, with standard errors clustered by author unless noted. Outcomes are $\log(1+\text{count})$ except in the any-engagement and Poisson rows. $^{a}$Also keeps the 6,226 posts with no author identifier. $^{*}P<0.05$, $^{**}P<0.01$, $^{***}P<0.001$.}
\label{tab:robustness}
\resizebox{\textwidth}{!}{%
\begin{tabular}{lrrrrrr}
\toprule
 & \multicolumn{3}{c}{\textbf{Across posts}} & \multicolumn{3}{c}{\textbf{Within author}} \\
\cmidrule(lr){2-4}\cmidrule(lr){5-7}
\textbf{Check} & Engagement & Reactions & Comments & Engagement & Reactions & Comments \\ \midrule
Baseline & $+0.085^{***}$ (0.021) & $-0.057^{**}$ (0.021) & $+0.953^{***}$ (0.021) & $+0.152^{***}$ (0.013) & $+0.146^{***}$ (0.013) & $+0.190^{***}$ (0.014) \\
Not clustered by author (HC1)$^{a}$ & $+0.090^{***}$ (0.012) & $-0.053^{***}$ (0.012) & $+0.952^{***}$ (0.012) & --- & --- & --- \\
Top 1\% most-engaged posts removed & $+0.054^{**}$ (0.020) & $-0.089^{***}$ (0.020) & $+0.931^{***}$ (0.020) & $+0.158^{***}$ (0.013) & $+0.153^{***}$ (0.012) & $+0.193^{***}$ (0.014) \\
Any engagement (yes/no, linear probability) & $0.000$ (0.002) & $0.000$ (0.002) & $+0.101^{***}$ (0.005) & $+0.003$ (0.002) & $+0.003$ (0.002) & $+0.029^{***}$ (0.005) \\
High-confidence topic labels only & $+0.073^{***}$ (0.021) & $-0.072^{***}$ (0.021) & $+0.974^{***}$ (0.021) & $+0.150^{***}$ (0.014) & $+0.147^{***}$ (0.014) & $+0.184^{***}$ (0.015) \\
Repeated posts left in & $+0.071^{**}$ (0.026) & $-0.083^{**}$ (0.025) & $+1.002^{***}$ (0.025) & $+0.169^{***}$ (0.016) & $+0.161^{***}$ (0.015) & $+0.227^{***}$ (0.017) \\
Poisson (PPML) on raw counts & $+0.499^{***}$ (0.054) & $+0.420^{***}$ (0.058) & $+1.138^{***}$ (0.060) & --- & --- & --- \\
\bottomrule
\end{tabular}}
\end{table}

\section*{Measures}

\begin{table}[H]
\centering
\small
\caption{\textbf{Requests for engagement and formatting features.} Detected with case-insensitive regular expressions on the post text.}
\label{tab:features_def}
\begin{tabular}{p{0.22\linewidth}p{0.72\linewidth}}
\toprule
Feature & Pattern \\
\midrule
Request to comment & The verb \textit{comment} (German \textit{kommentier-}, \textit{schreib-}; Dutch \textit{reageer}; Spanish \textit{comenta}, \textit{escribe}; French \textit{commente}; Portuguese \textit{comente}) followed by a quoted keyword; \textit{comment below}; \textit{drop a \ldots\ in the comments} or \textit{below}; \textit{let me know in the comments}; \textit{comment interested}; \textit{type} or \textit{write} a quoted word \textit{below}; \textit{in die Kommentare}, \textit{in de comments}, \textit{en los comentarios}, \textit{en commentaire} \\[3pt]
Request to share or repost & \textit{repost}, \textit{reshare}, \textit{share this}, \textit{share with your network}; German \textit{teilen}; Spanish and Portuguese \textit{compartir}, \textit{comparte} \\[3pt]
Request to follow or connect & \textit{follow me}, \textit{follow us}, \textit{follow for more}, \textit{hit the} bell, \textit{connect with me}, \textit{send me a connection}; German \textit{vernetz-} \\[3pt]
Question at the end & A question mark in the closing line, ignoring links and lines of only hashtags or emoji; for a post written as one paragraph, in its last sentence (yes/no) \\[3pt]
Hashtag & A \# followed by a letter or digit (yes/no) \\[3pt]
Link & \textit{http://}, \textit{https://}, or \textit{www.} in the post (yes/no) \\[3pt]
Emoji & Number of characters in the Unicode ranges U+1F300--U+1FAFF and U+2600--U+27BF; entered as $\log(1+n)$ \\[3pt]
Bulleted lines & Number of lines that begin with a bullet, dash, arrow, check mark, pointing hand, diamond, or star symbol, or with a number followed by a period, parenthesis, or slash; entered as $\log(1+n)$ \\
\bottomrule
\end{tabular}
\end{table}

\begin{table}[H]
\centering
\small
\caption{\textbf{Post characteristics in Fig.~\ref{fig:authors}\textit{A} and Fig.~\ref{fig:post_characteristics}.} Each is computed per post and, in Fig.~\ref{fig:authors}\textit{A}, averaged within author. Requests for engagement are defined in Table~\ref{tab:features_def}.}
\label{tab:traits_def}
\begin{tabular}{p{0.21\linewidth}p{0.72\linewidth}}
\toprule
Characteristic & Definition \\
\midrule
Words & Number of whitespace-delimited words \\
Average word length & Non-whitespace characters divided by the number of words \\
Link & Text contains \texttt{http://}, \texttt{https://} or \texttt{www.} \\
Digits & Digits per 100 words \\
Capital letters & Share of Latin letters that are upper case \\
@-mention & Text contains @ followed by a letter or digit \\
Exclamation mark & Text contains at least one exclamation mark \\
Emoji & Emoji (as in Table~\ref{tab:features_def}) per 100 words \\
Ending with a question & The post ends with a question, as in Table~\ref{tab:features_def} (yes/no) \\
Sentiment & $P(\text{positive})-P(\text{negative})$ from \texttt{cardiffnlp/twitter-xlm-roberta-base-sentiment}, from $-1$ to $+1$ \\
Reshared & The platform marks the post as republished \\
\bottomrule
\end{tabular}
\end{table}

\begin{table}[H]
\centering
\small
\setlength{\tabcolsep}{4pt}
\caption{\textbf{AI scores by country.} Data for Fig.~\ref{fig:authors}\textit{B}, for each country listed on the profile of at least 150 authors with at least two posts. Posts were classified as English with \texttt{langdetect} (Methods), and the mean over English-language posts is over the authors with at least one such post. The last two columns give the share of the country's authors in the quartiles with the lowest and the highest AI scores among all 106,968 authors with at least two posts. Standard errors are in parentheses.}
\label{tab:country_language}
\begin{tabular}{lrrrrrrr}
\toprule
 & \multicolumn{2}{c}{All posts} & \multicolumn{3}{c}{English-language posts} & \multicolumn{2}{c}{Quartile of AI score} \\
\cmidrule(lr){2-3}\cmidrule(lr){4-6}\cmidrule(lr){7-8}
Country & Authors & Mean AI score & Share of posts & Authors & Mean AI score & Least & Most \\
\midrule
Brazil & 942 & 0.589 (0.003) & 5.7\% & 118 & 0.483 (0.009) & 6.7\% & 56.9\% \\
Italy & 194 & 0.548 (0.008) & 40.2\% & 96 & 0.501 (0.011) & 16.5\% & 40.2\% \\
Bangladesh & 327 & 0.545 (0.004) & 96.3\% & 324 & 0.549 (0.005) & 8.9\% & 45.9\% \\
Netherlands & 279 & 0.535 (0.006) & 80.8\% & 228 & 0.510 (0.006) & 15.8\% & 36.2\% \\
Germany & 576 & 0.528 (0.004) & 64.0\% & 419 & 0.506 (0.004) & 16.8\% & 37.5\% \\
Nigeria & 559 & 0.521 (0.003) & 99.8\% & 559 & 0.521 (0.003) & 11.4\% & 29.9\% \\
Pakistan & 3{,}635 & 0.519 (0.001) & 98.6\% & 3{,}627 & 0.518 (0.001) & 12.5\% & 29.6\% \\
Saudi Arabia & 306 & 0.518 (0.005) & 78.0\% & 277 & 0.514 (0.006) & 14.1\% & 31.7\% \\
India & 3{,}085 & 0.513 (0.001) & 99.6\% & 3{,}085 & 0.513 (0.001) & 15.5\% & 28.1\% \\
United Arab Emirates & 1{,}611 & 0.513 (0.002) & 95.9\% & 1{,}580 & 0.512 (0.002) & 17.2\% & 29.7\% \\
Spain & 340 & 0.510 (0.005) & 51.6\% & 191 & 0.496 (0.006) & 17.6\% & 26.2\% \\
Switzerland & 254 & 0.505 (0.006) & 86.2\% & 229 & 0.492 (0.006) & 23.2\% & 26.8\% \\
Canada & 748 & 0.503 (0.004) & 98.4\% & 740 & 0.502 (0.004) & 22.3\% & 27.8\% \\
Australia & 375 & 0.498 (0.005) & 99.9\% & 375 & 0.498 (0.005) & 22.1\% & 25.6\% \\
Chile & 159 & 0.495 (0.006) & 5.4\% & 18 & 0.497 (0.021) & 22.0\% & 20.1\% \\
United States & 6{,}554 & 0.492 (0.001) & 99.3\% & 6{,}519 & 0.492 (0.001) & 26.5\% & 24.0\% \\
Colombia & 339 & 0.490 (0.006) & 8.6\% & 44 & 0.444 (0.015) & 26.3\% & 24.2\% \\
France & 614 & 0.482 (0.004) & 48.5\% & 336 & 0.467 (0.005) & 36.5\% & 21.3\% \\
United Kingdom & 2{,}788 & 0.480 (0.002) & 99.6\% & 2{,}785 & 0.480 (0.002) & 30.8\% & 19.8\% \\
Singapore & 274 & 0.475 (0.005) & 99.8\% & 274 & 0.475 (0.005) & 34.3\% & 16.4\% \\
\midrule
All who listed a country & 26{,}976 & 0.507 & 89.7\% & 24{,}320 & 0.501 & 20.6\% & 27.6\% \\
\bottomrule
\end{tabular}
\end{table}

\begin{table}[H]
\centering
\small
\caption{\textbf{Topic categories.} Definitions are those given to \texttt{gpt-5.6-luna} (low reasoning effort; OpenAI Batch API, 10 posts per request, each post cut at 12,000 characters). The prompt also told the model to ignore instructions written inside a post, and how to separate labour from economy, business and finance, human interest from society, and science and technology from business. The model returned one primary topic per post from this list and its confidence in it: high for 89.39\% of posts, medium for 10.58\%, low for 0.03\%.}
\label{tab:topics_def}
\begin{tabular}{p{0.3\linewidth}p{0.63\linewidth}}
\toprule
Topic & Definition given to the model \\
\midrule
Economy, business and finance & Text about companies, products and services, industries, national economies, international trade, banks, currencies (including cryptocurrencies), business and trade associations, economic trends and indicators (e.g., inflation, employment statistics, GDP, and mortgages), international economic institutions, and utilities (e.g., electricity, heating, waste management, and water supply). \\
Labour & Text about employment, employment legislation, employees and employers, commuting, parental leave, volunteering, wages, social security, the labour market, retirement, unemployment, and unions. \\
Science and technology & Text about the natural and social sciences, mathematics, technology and engineering, scientific institutions, research, scientific publications, and innovation. \\
Education & Text about formal or informal learning, including schools, curricula, grading, remote learning, teachers, and students. \\
Human interest & Text about the lives and behaviour of royalty and celebrities, people receiving awards, ceremonies (e.g., graduations, weddings, funerals, and launch celebrations), birthdays, anniversaries, and amusing or foolish mistakes. \\
Society & Text about social interactions (e.g., networking), demographic analyses and censuses, discrimination, inclusion and equity, migration, communities and minority groups (e.g., LGBTQ people, older people, children, and Indigenous peoples), homelessness, poverty, social problems (e.g., addiction and bullying), ethical issues (e.g., suicide, euthanasia, and sexual behaviour), social services and charity, relationships (e.g., dating, divorce, and marriage), and family life (e.g., family planning, adoption, abortion, contraception, pregnancy, and parenting). \\
Health & Text about diseases, injuries, mental health problems, treatments, diets, vaccines, drugs, government healthcare, hospitals, medical staff, and health insurance. \\
Arts, culture, entertainment and media & Text about cinema, dance, fashion, hairstyles, jewellery, festivals, literature, music, theatre, television shows, painting, photography, woodworking, art exhibitions, libraries and museums, language, cultural heritage, news media, radio and television, social media, influencers, and disinformation. \\
Politics & Text about the exercise of power at the local, regional, national, or international level, including elections, fundamental rights, governments, non-governmental organisations, political crises, nonviolent international relations, public employees, and government policies. \\
Lifestyle and leisure & Text about hobbies, clubs and societies, games, lotteries, food and drink interests, car and motorcycle enthusiasts, public holidays, leisure venues (e.g., amusement parks, cafés, bars, and restaurants), exercise and fitness, outdoor recreation (e.g., fishing and hunting), travel and tourism, mental well-being, parties, and the upkeep and decoration of homes and gardens. \\
Environment & Text about climate change, energy conservation, sustainability, pollution, population growth, natural resources, forests, mountains, bodies of water, ecosystems, animals, flowers, and other plants. \\
Crime, law and justice & Text about crimes and other illegal activities, courts, law, and law enforcement, including judges, lawyers, trials, and the punishment of offenders. \\
Sport & Text about competitive sports (e.g., basketball, football, swimming, athletics, chess, dog racing, diving, golf, gymnastics, martial arts, and climbing), sporting achievements and events, sports organisations and clubs, venues (e.g., stadiums and gymnasiums), referees, coaches, and drug use in sport. \\
Religion & Text about religions, cults, religious conflicts, relations between religion and government, churches, religious holidays and festivals, religious leaders, rituals, and religious texts. \\
Conflict, war and peace & Text about terrorism, wars, victims of war, cyberwarfare, civil unrest (e.g., demonstrations, riots, and rebellions), peace talks, and other peace activities. \\
Disaster, accident and emergency incident & Text about human-caused or natural events resulting in injury, death, or damage (e.g., explosions, transport accidents, famine, drowning, and natural disasters), as well as emergency planning and response. \\
Weather & Text about weather forecasts, weather phenomena, and weather warnings. \\
\bottomrule
\end{tabular}
\end{table}

\section*{Validation of the AI score}

\begin{table}[H]
\centering
\small
\caption{\textbf{Surface markers of LLM text and the AI score.} Each marker is regressed on the AI score (0 to 1) and log word count, across posts without fixed effects and within authors with author fixed effects. Standard errors, clustered by author, are in parentheses. Em-dashes and LLM-typical words are counted per 100 words. The ``not just X, but Y'' construction (coded yes or no) is \textit{not just}, \textit{not only}, or \textit{not merely} followed within 80 characters by \textit{but} or \textit{it's}, or \textit{isn't just} or \textit{isn't about} followed within 60 characters by \textit{it's}. The LLM-typical words are based on words reported as overused by LLMs \protect\cite{kobak2025delving,reinhart2025llms}: \textit{delve}, \textit{tapestry}, \textit{elevate}, \textit{leverage}, \textit{seamless}, \textit{robust}, \textit{landscape}, \textit{navigate}, \textit{empower}, \textit{foster}, \textit{unlock}, \textit{game changer}; for \textit{leverage}, \textit{seamless}, \textit{navigate}, \textit{empower}, \textit{foster} and \textit{unlock}, other forms of the word count too (e.g., \textit{leveraging}). Emoji-led lines are $\log(1+n)$ of lines that begin with an emoji. $^{*}P<0.05$, $^{**}P<0.01$, $^{***}P<0.001$.}
\label{tab:markers}
\begin{tabular}{lrr}
\toprule
Marker & Across posts & Within author \\
\midrule
Em-dashes per 100 words & $+0.259^{***}$ & $+0.137^{***}$ \\
 & (0.009) & (0.011) \\[2pt]
``Not just X, but Y'' & $-0.001$ & $+0.009^{***}$ \\
 & (0.002) & (0.003) \\[2pt]
LLM-typical words per 100 words & $-0.007^{*}$ & $+0.031^{***}$ \\
 & (0.003) & (0.005) \\[2pt]
Emoji-led lines, $\log(1+n)$ & $+0.350^{***}$ & $+0.165^{***}$ \\
 & (0.009) & (0.009) \\
\midrule
Posts & 872,469 & 559,149 \\
Authors & 420,288 & 106,968 \\
\bottomrule
\end{tabular}
\end{table}

\section*{Change within authors over time}
\begin{figure}[H]
\centering
\includegraphics[width=\textwidth]{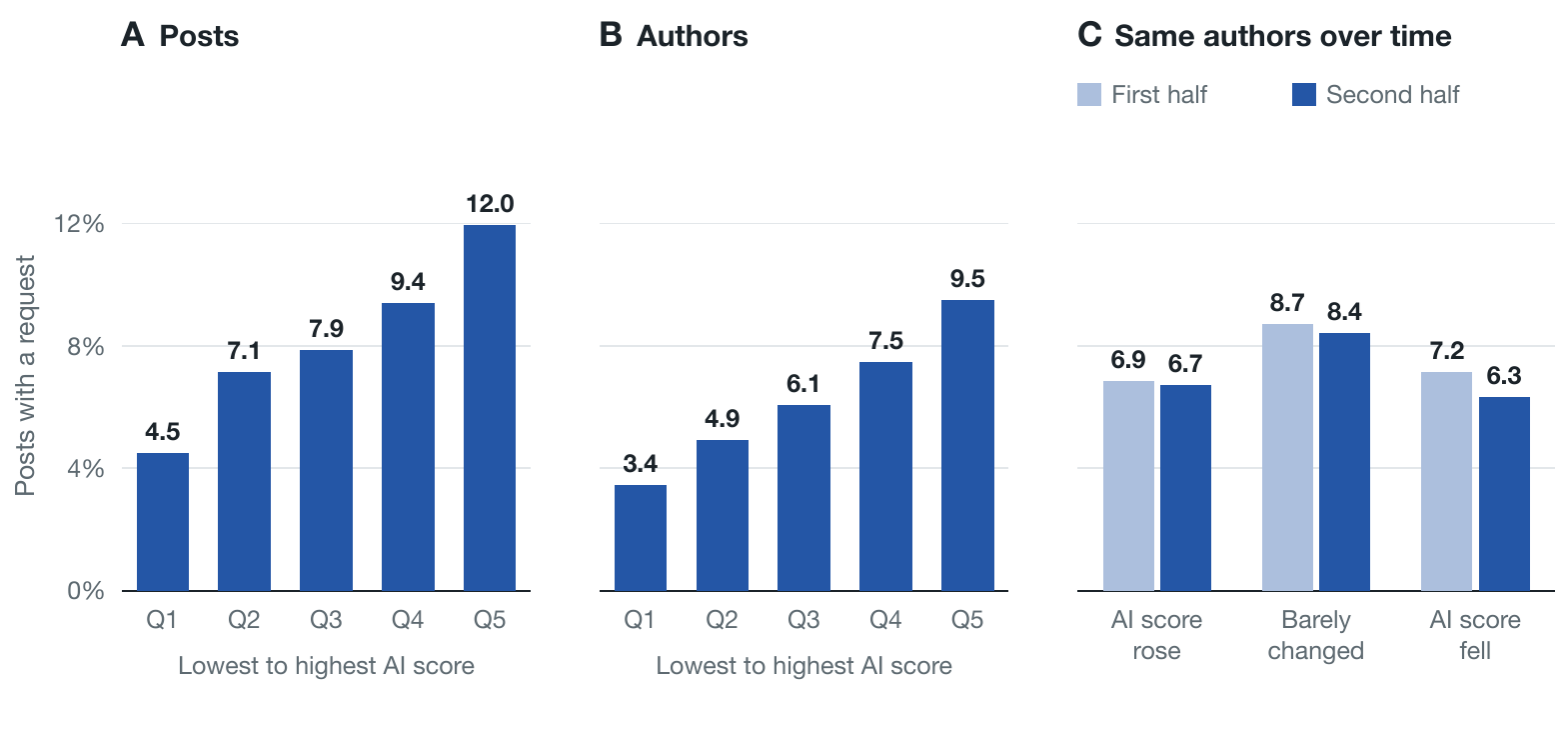}
\caption{\textbf{Requests for engagement in posts, across authors, and within the same authors over time.} Share of posts that asked readers to comment, share, or follow, on a common scale. (\textit{A}) All 995,629 posts, by fifth of the post's AI score. (\textit{B}) The 106,968 authors with at least two posts, by fifth of the mean AI score of their posts, as in Fig.~\ref{fig:authors}\textit{A}. Bars are the average of the authors' own shares. (\textit{C}) Authors with at least four posts, in the first and second half of their own posting window, grouped as in Fig.~\ref{fig:adoption}. The fifth with the highest AI scores asked more often than the fifth with the lowest, by 7.5 percentage points among posts and 6.1 among authors (both $P<0.001$). Relative to authors whose AI scores barely changed, authors whose scores rose did not ask more often ($+0.2$ percentage points, Welch $t$ test, $P=0.58$).}
\label{fig:requests_three}
\end{figure}

We then compared these changes between the groups. Compared with authors whose scores barely changed, authors whose scores fell lost a further 8.1\% in engagement ($t(10{,}793)=-5.5$, $P<0.001$) and 11.8\% in comments ($t(10{,}247)=-7.0$, $P<0.001$), and authors whose scores rose gained a further 3.7\% in engagement ($t(11{,}540)=2.4$, $P=0.016$) and 6.7\% in comments ($t(10{,}950)=3.7$, $P<0.001$). Compared with authors whose scores fell, authors whose scores rose gained 12.8\% more engagement ($t(12{,}887)=6.6$, $P<0.001$) and 20.9\% more comments ($t(12{,}888)=8.7$, $P<0.001$). Authors whose AI scores rose wrote 6.8\% more words per post in the second half ($t(6{,}615)=11.1$, $P<0.001$), whereas authors whose scores fell wrote 4.1\% fewer ($t(6{,}313)=-6.9$, $P<0.001$). Compared with authors whose scores barely changed ($+1.3\%$), these were differences of $+5.4\%$ ($t(11{,}206)=7.6$, $P<0.001$) and $-5.3\%$ ($t(10{,}570)=-7.8$, $P<0.001$). The authors whose AI scores rose asked readers to comment, share, or follow about as often as before, in 6.9\% of their posts in the first half and 6.7\% in the second ($+0.2$ percentage points relative to authors whose scores barely changed, $t(11{,}985)=0.6$, $P=0.58$; Fig.~\ref{fig:requests_three}\textit{C}).

\begin{table}[H]
\centering
\small
\caption{\textbf{Average within-author change by group, authors with at least four posts.} The first three columns are each group's average of its authors' changes from the first to the second half, the mean change in $\log(1+\text{value})$ shown as a percentage. The last three columns give the difference between two groups in that mean change (Welch $t$ test), and under it the raw difference. Groups: mean AI score up by at least 0.1 (rose), down by at least 0.1 (fell), or within 0.02 (barely changed). Engagement, reactions, and comments are net of calendar-week effects (Materials and Methods). $^{*}P<0.05$, $^{**}P<0.01$, $^{***}P<0.001$.}
\label{tab:adoption_groups}
\resizebox{\textwidth}{!}{%
\begin{tabular}{lrrr|rrr}
\toprule
 & \multicolumn{3}{c|}{Own change} & \multicolumn{3}{c}{Difference} \\
Change in & Rose $\geq$0.1 & Changed $<$0.02 & Fell $\geq$0.1 & Rose vs barely changed & Fell vs barely changed & Rose vs fell \\
\midrule
Engagement per post & $+5.6\%$ & $+1.9\%$ & $-6.4\%$ & $+3.7\%^{*}$ & $-8.1\%^{***}$ & $+12.8\%^{***}$ \\
 & & & & \footnotesize{$+0.036$} & \footnotesize{$-0.085$} & \footnotesize{$+0.121$} \\[2pt]
Reactions per post & $+5.4\%$ & $+2.0\%$ & $-5.6\%$ & $+3.4\%^{*}$ & $-7.4\%^{***}$ & $+11.7\%^{***}$ \\
 & & & & \footnotesize{$+0.033$} & \footnotesize{$-0.077$} & \footnotesize{$+0.110$} \\[2pt]
Comments per post & $+7.9\%$ & $+1.2\%$ & $-10.7\%$ & $+6.7\%^{***}$ & $-11.8\%^{***}$ & $+20.9\%^{***}$ \\
 & & & & \footnotesize{$+0.065$} & \footnotesize{$-0.125$} & \footnotesize{$+0.190$} \\[2pt]
\midrule
Words per post & $+6.8\%$ & $+1.3\%$ & $-4.1\%$ & $+5.4\%^{***}$ & $-5.3\%^{***}$ & $+11.3\%^{***}$ \\
 & & & & \footnotesize{$+0.053$} & \footnotesize{$-0.055$} & \footnotesize{$+0.107$} \\[2pt]
Posts per half & $+10.7\%$ & $-5.7\%$ & $+4.5\%$ & $+17.4\%^{***}$ & $+10.8\%^{***}$ & $+5.9\%^{***}$ \\
 & & & & \footnotesize{$+0.160$} & \footnotesize{$+0.102$} & \footnotesize{$+0.058$} \\[2pt]
Total words written & $+22.8\%$ & $-6.8\%$ & $+2.7\%$ & $+31.8\%^{***}$ & $+10.1\%^{***}$ & $+19.7\%^{***}$ \\
 & & & & \footnotesize{$+0.276$} & \footnotesize{$+0.097$} & \footnotesize{$+0.179$} \\[2pt]
\bottomrule
\end{tabular}}
\end{table}

\begin{table}[H]
\centering
\small
\caption{\textbf{Change within authors as their AI scores rise.} Increases in an author's mean AI score were associated with more engagement per post, especially comments, and with more writing. Authors with at least four posts were split at the midpoint of their own posting window, and each quantity is the change from the first to the second half. Cells show the percentage change in the outcome for a 0.1 increase in the author's mean AI score, with the coefficient on the 0--1 score and its HC3 standard error below. The second column adds the changes in log word count, sentiment, and the number of posts in the half, except where one of these is the outcome. Engagement, reactions, and comments are net of calendar-week effects (Materials and Methods). Data are from 44,936 authors, and the SD of the change in the AI score is 0.105. $^{*}P<0.05$, $^{**}P<0.01$, $^{***}P<0.001$.}
\label{tab:adoption}
\begin{tabular}{lrr}
\toprule
Change in & AI change & + controls \\
\midrule
Engagement per post & $+3.8\%^{***}$ & $+2.4\%^{***}$ \\
 & \footnotesize{$+0.378$ (0.047)} & \footnotesize{$+0.235$ (0.046)} \\[2pt]
Reactions per post & $+3.6\%^{***}$ & $+2.1\%^{***}$ \\
 & \footnotesize{$+0.350$ (0.046)} & \footnotesize{$+0.212$ (0.045)} \\[2pt]
Comments per post & $+6.1\%^{***}$ & $+4.4\%^{***}$ \\
 & \footnotesize{$+0.593$ (0.058)} & \footnotesize{$+0.433$ (0.057)} \\[2pt]
\midrule
Words per post & $+3.1\%^{***}$ & --- \\
 & \footnotesize{$+0.310$ (0.022)} &  \\[2pt]
Posts per half & $+1.7\%^{***}$ & --- \\
 & \footnotesize{$+0.167$ (0.035)} &  \\[2pt]
Total words written & $+5.3\%^{***}$ & --- \\
 & \footnotesize{$+0.519$ (0.055)} &  \\[2pt]
\bottomrule
\end{tabular}
\end{table}

\begin{figure}[H]
\centering
\includegraphics[width=\textwidth]{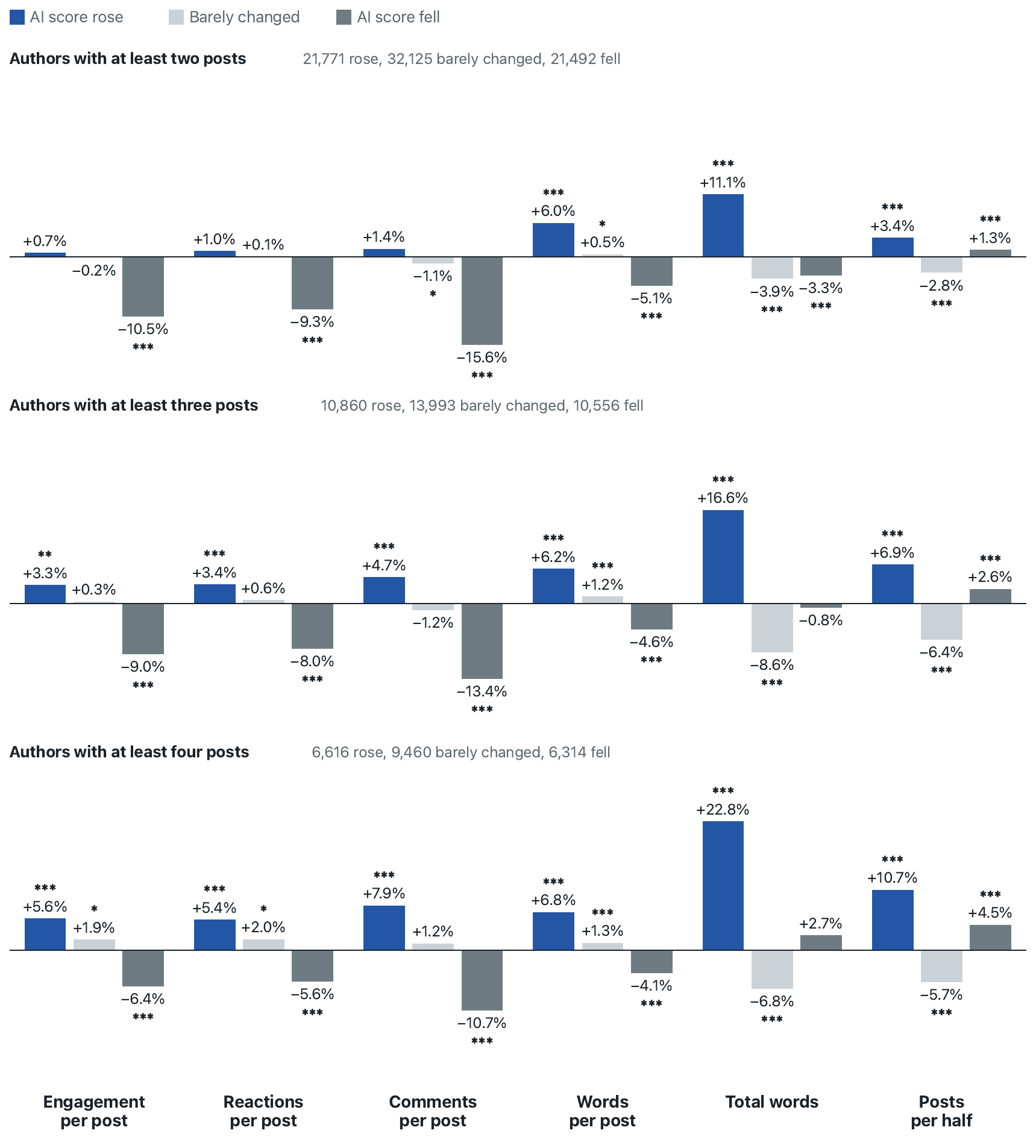}
\caption{\textbf{The over-time pattern in three samples.} As Fig.~\ref{fig:adoption} (authors with at least four posts, bottom row), with authors with at least two and three posts for comparison, on a common scale.}
\label{fig:S7}
\end{figure}

We repeated these analyses with looser and stricter samples: authors with at least two posts and with at least three posts, each with at least one post in each half, and authors with at least four posts and at least two in each half. Relative to authors whose scores barely changed, authors whose AI scores fell lost 10.4\% of their engagement in the first of these samples ($t(32{,}629)=-12.7$, $P<0.001$), 9.3\% in the second ($t(18{,}240)=-7.9$, $P<0.001$), and 9.7\% in the third ($t(6{,}203)=-5.9$, $P<0.001$). A 0.1 rise in the AI score was associated with 3.1\% more engagement ($t(123{,}164)=12.0$, $P<0.001$), 3.6\% more ($t(66{,}130)=9.5$, $P<0.001$), and 4.4\% more ($t(31{,}883)=7.6$, $P<0.001$; Fig.~\ref{fig:S7} and Table~\ref{tab:adoption_samples}). The gain for authors whose scores rose was smaller and less consistent: $+0.9\%$ ($t(33{,}556)=1.1$, $P=0.29$), $+2.9\%$ ($t(19{,}101)=2.4$, $P=0.017$), and $+4.6\%$ ($t(6{,}870)=2.7$, $P=0.008$).

\begin{table}[H]
\centering
\small
\caption{\textbf{The over-time association in four samples.} Table~\ref{tab:adoption} repeated for four samples. Cells are the percentage change in the outcome for a 0.1 increase in the author's mean AI score. Rows marked yes add the controls of Table~\ref{tab:adoption}. The group rows are the differences of Table~\ref{tab:adoption_groups} for engagement per post. $^{*}P<0.05$, $^{**}P<0.01$, $^{***}P<0.001$.}
\label{tab:adoption_samples}
\begin{tabular}{llrrrr}
\toprule
Change in & Controls & 4+ posts (main) & 2+ posts & 3+ posts & 4+ posts, 2+ per half \\
\midrule
Engagement per post & none & $+3.8\%^{***}$ & $+3.1\%^{***}$ & $+3.6\%^{***}$ & $+4.4\%^{***}$ \\
 & yes & $+2.4\%^{***}$ & $+1.9\%^{***}$ & $+2.2\%^{***}$ & $+3.0\%^{***}$ \\
Reactions per post & none & $+3.6\%^{***}$ & $+2.9\%^{***}$ & $+3.3\%^{***}$ & $+4.1\%^{***}$ \\
 & yes & $+2.1\%^{***}$ & $+1.7\%^{***}$ & $+2.0\%^{***}$ & $+2.8\%^{***}$ \\
Comments per post & none & $+6.1\%^{***}$ & $+4.9\%^{***}$ & $+5.5\%^{***}$ & $+6.3\%^{***}$ \\
 & yes & $+4.4\%^{***}$ & $+3.5\%^{***}$ & $+3.9\%^{***}$ & $+4.7\%^{***}$ \\
\midrule
Words per post & none & $+3.1\%^{***}$ & $+2.9\%^{***}$ & $+3.1\%^{***}$ & $+3.4\%^{***}$ \\
Posts per half & none & $+1.7\%^{***}$ & $+0.5\%^{***}$ & $+1.2\%^{***}$ & $+1.4\%^{***}$ \\
Total words written & none & $+5.3\%^{***}$ & $+3.7\%^{***}$ & $+4.8\%^{***}$ & $+5.0\%^{***}$ \\
\midrule
\multicolumn{6}{l}{Difference between groups in the change in engagement per post} \\
Fell vs barely changed & & $-8.1\%^{***}$ & $-10.4\%^{***}$ & $-9.3\%^{***}$ & $-9.7\%^{***}$ \\
Rose vs barely changed & & $+3.7\%^{*}$ & $+0.9\%$ & $+2.9\%^{*}$ & $+4.6\%^{**}$ \\
\midrule
Authors & & 44,936 & 123,166 & 66,132 & 31,885 \\
\bottomrule
\end{tabular}
\end{table}

\section*{Common features and mediation}

Posts with higher AI scores contained more of each request, and requests predicted comments, so we decomposed the AI--comments association accordingly (Methods). Within authors, the ten features accounted for $+0.137$ of a $+0.189$ association with comments (95\% CI of the indirect effect $+0.131$ to $+0.143$; Table~\ref{tab:mediation_within}), of which post length contributed $+0.085$, links $+0.048$, and the three requests and a closing question $+0.007$. For engagement, they accounted for $+0.140$ of $+0.152$ ($+0.134$ to $+0.147$), of which post length contributed $+0.091$ and links $+0.052$. Across posts, they accounted for $+0.608$ of the $+0.953$ association with comments ($+0.591$ to $+0.624$; Table~\ref{tab:mediation}), of which post length contributed $+0.255$, links $+0.234$, and the three requests and a closing question $+0.106$.

\begin{table}[H]
\centering
\small
\caption{\textbf{The ten features across posts and within authors.} Within author, the multilevel model of Table~\ref{tab:tactics}. Across posts, the same features and controls in an ordinary least-squares regression on all posts. Outcomes are $\log(1+\text{count})$. Words, bulleted lines, and emoji are logged counts, $\log(\text{words})$ and $\log(1+n)$; for bulleted lines and emoji, the share is of posts with at least one. Sentiment is $P(\text{positive})-P(\text{negative})$, from $-1$ to $+1$. The share of posts is out of the 559,149 posts in the within-author model. Standard errors, clustered by author, are in parentheses. $^{*}P<0.05$, $^{**}P<0.01$, $^{***}P<0.001$.}
\label{tab:tactics_full}
\begin{tabular}{lrrrrrrr}
\toprule
 & & \multicolumn{3}{c}{Across posts} & \multicolumn{3}{c}{Within author} \\
\cmidrule(lr){3-5}\cmidrule(lr){6-8}
Feature & \% of posts & Engagement & Reactions & Comments & Engagement & Reactions & Comments \\
\midrule
Words & --- & $+0.58^{***}$ & $+0.57^{***}$ & $+0.52^{***}$ & $+0.31^{***}$ & $+0.30^{***}$ & $+0.29^{***}$ \\
 & & (0.007) & (0.007) & (0.006) & (0.004) & (0.004) & (0.005) \\[2pt]
Bulleted lines & 33.2 & $-0.09^{***}$ & $-0.11^{***}$ & $-0.01^{***}$ & $-0.04^{***}$ & $-0.04^{***}$ & $-0.03^{***}$ \\
 & & (0.004) & (0.004) & (0.004) & (0.002) & (0.002) & (0.002) \\[2pt]
Emoji & 48.8 & $-0.10^{***}$ & $-0.10^{***}$ & $-0.05^{***}$ & $-0.01^{*}$ & $-0.01^{**}$ & $0.00$ \\
 & & (0.005) & (0.005) & (0.004) & (0.003) & (0.002) & (0.003) \\[2pt]
Requests to comment & 3.1 & $+0.17^{***}$ & $-0.10^{***}$ & $+0.82^{***}$ & $+0.21^{***}$ & $+0.08^{***}$ & $+0.47^{***}$ \\
 & & (0.021) & (0.017) & (0.027) & (0.015) & (0.011) & (0.021) \\[2pt]
Requests to share & 5.6 & $+0.38^{***}$ & $+0.32^{***}$ & $+0.53^{***}$ & $+0.09^{***}$ & $+0.08^{***}$ & $+0.10^{***}$ \\
 & & (0.019) & (0.019) & (0.020) & (0.009) & (0.009) & (0.010) \\[2pt]
Requests to follow & 2.4 & $+0.31^{***}$ & $+0.19^{***}$ & $+0.55^{***}$ & $+0.09^{***}$ & $+0.05^{***}$ & $+0.12^{***}$ \\
 & & (0.031) & (0.031) & (0.032) & (0.016) & (0.014) & (0.018) \\[2pt]
Ending with a question & 9.7 & $-0.24^{***}$ & $-0.28^{***}$ & $-0.01$ & $-0.06^{***}$ & $-0.06^{***}$ & $-0.01^{*}$ \\
 & & (0.011) & (0.011) & (0.011) & (0.005) & (0.005) & (0.005) \\[2pt]
Hashtags & 48.7 & $-0.35^{***}$ & $-0.31^{***}$ & $-0.46^{***}$ & $-0.04^{***}$ & $-0.04^{***}$ & $-0.08^{***}$ \\
 & & (0.008) & (0.008) & (0.007) & (0.005) & (0.005) & (0.005) \\[2pt]
Links & 23.0 & $-0.20^{***}$ & $-0.15^{***}$ & $-0.45^{***}$ & $-0.19^{***}$ & $-0.18^{***}$ & $-0.18^{***}$ \\
 & & (0.010) & (0.010) & (0.008) & (0.005) & (0.005) & (0.005) \\[2pt]
Sentiment & --- & $+0.47^{***}$ & $+0.50^{***}$ & $+0.14^{***}$ & $+0.30^{***}$ & $+0.31^{***}$ & $+0.18^{***}$ \\
 & & (0.010) & (0.010) & (0.009) & (0.006) & (0.006) & (0.007) \\[2pt]
AI score (0--1) & --- & $-0.24^{***}$ & $-0.30^{***}$ & $+0.35^{***}$ & $+0.01$ & $+0.01$ & $+0.05^{***}$ \\
 & & (0.020) & (0.020) & (0.019) & (0.013) & (0.012) & (0.014) \\[2pt]
\midrule
Posts / authors & & \multicolumn{3}{c}{872,469 / 420,288} & \multicolumn{3}{c}{559,149 / 106,968} \\
\bottomrule
\end{tabular}
\end{table}

\begin{table}[H]
\centering
\small
\setlength{\tabcolsep}{4pt}
\caption{\textbf{How much of the association with the AI score runs through the ten features, across posts.} The features carried 63.8\% of the association with comments, most of it through post length and links. The change in each feature is for an AI score going from 0 to 1, and its association with comments holds the AI score constant (across posts in Table~\ref{tab:tactics_full}). For the indirect effects by group, the percentage is the group's share of the total effect. The 95\% intervals of the indirect effects are from 1,000 bootstrap samples of authors ($^{\dagger}$interval excludes 0). Standard errors are clustered by author. The within-author decomposition is in Table~\ref{tab:mediation_within}. $^{*}P<0.05$, $^{**}P<0.01$, $^{***}P<0.001$.}
\label{tab:mediation}
\begin{tabular}{lrrl}
\toprule
 & \multicolumn{1}{c}{Change in the feature} & \multicolumn{1}{c}{Feature's association} & \multicolumn{1}{c}{Indirect effect on comments} \\
Feature & \multicolumn{1}{c}{as the AI score rises ($a$)} & \multicolumn{1}{c}{with comments ($b$)} & \multicolumn{1}{c}{($a\times b$) [95\% CI]} \\
\midrule
Words & $+0.491^{***}$ & $+0.52^{***}$ & $+0.2551^{\dagger}$ \footnotesize{[+0.2450, +0.2637]} \\
Bulleted lines & $+0.609^{***}$ & $-0.01^{***}$ & $-0.0085^{\dagger}$ \footnotesize{[-0.0133, -0.0033]} \\
Emoji & $+0.386^{***}$ & $-0.05^{***}$ & $-0.0208^{\dagger}$ \footnotesize{[-0.0241, -0.0176]} \\
Requests to comment & $+0.054^{***}$ & $+0.82^{***}$ & $+0.0449^{\dagger}$ \footnotesize{[+0.0406, +0.0494]} \\
Requests to share & $+0.082^{***}$ & $+0.53^{***}$ & $+0.0434^{\dagger}$ \footnotesize{[+0.0380, +0.0487]} \\
Requests to follow & $+0.033^{***}$ & $+0.55^{***}$ & $+0.0182^{\dagger}$ \footnotesize{[+0.0149, +0.0220]} \\
Ending with a question & $+0.078^{***}$ & $-0.01$ & $-0.0006$ \footnotesize{[-0.0025, +0.0010]} \\
Hashtags & $-0.118^{***}$ & $-0.46^{***}$ & $+0.0542^{\dagger}$ \footnotesize{[+0.0480, +0.0605]} \\
Links & $-0.525^{***}$ & $-0.45^{***}$ & $+0.2344^{\dagger}$ \footnotesize{[+0.2236, +0.2434]} \\
Sentiment & $-0.091^{***}$ & $+0.14^{***}$ & $-0.0124^{\dagger}$ \footnotesize{[-0.0144, -0.0103]} \\
\midrule
Indirect, post length & & $26.8\%$ & $+0.2551^{\dagger}$ \footnotesize{[+0.2450, +0.2637]} \\
Indirect, links & & $24.6\%$ & $+0.2344^{\dagger}$ \footnotesize{[+0.2236, +0.2434]} \\
Indirect, the four requests & & $11.1\%$ & $+0.1059^{\dagger}$ \footnotesize{[+0.0977, +0.1139]} \\
Indirect, the three formatting features & & $2.6\%$ & $+0.0249^{\dagger}$ \footnotesize{[+0.0166, +0.0328]} \\
Indirect, sentiment & & $-1.3\%$ & $-0.0124^{\dagger}$ \footnotesize{[-0.0144, -0.0103]} \\
Indirect, all ten features & &  & $+0.6079^{\dagger}$ \footnotesize{[+0.5907, +0.6243]} \\
Direct effect of the AI score ($c'$) & &  & $+0.346^{***}$ \footnotesize{(0.019)} \\
Total effect of the AI score ($c$) & &  & $+0.953^{***}$ \footnotesize{(0.021)} \\
Share of total through features & &  & 63.8\% \\
\midrule
Posts / authors & \multicolumn{3}{l}{872,469 / 420,288} \\
\bottomrule
\end{tabular}
\end{table}

\begin{table}[H]
\centering
\small
\setlength{\tabcolsep}{4pt}
\caption{\textbf{Mediation through the ten features within authors.} Compared with the same author's other posts, posts with higher AI scores were longer, less often contained a link, more often asked readers to comment, share, or follow, and less often ended with a question (first column). The features carried 92.4\% of the association with engagement and 72.4\% of the association with comments, most of it through post length and links. For each outcome, $b$ is the feature's association with the outcome, holding the AI score and the other features constant, and $a\times b$ is the indirect effect. For the indirect effects by group, the percentage is the group's share of the total effect. As in Table~\ref{tab:mediation}, from the multilevel model of Table~\ref{tab:tactics}. $^{*}P<0.05$, $^{**}P<0.01$, $^{***}P<0.001$.}
\label{tab:mediation_within}
\resizebox{\textwidth}{!}{%
\begin{tabular}{lrrlrl}
\toprule
 & \multicolumn{1}{c}{Change in the feature} & \multicolumn{2}{c}{Engagement} & \multicolumn{2}{c}{Comments} \\
\cmidrule(lr){3-4}\cmidrule(lr){5-6}
Feature & \multicolumn{1}{c}{as the AI score rises ($a$)} & \multicolumn{1}{c}{$b$} & \multicolumn{1}{c}{$a\times b$ [95\% CI]} & \multicolumn{1}{c}{$b$} & \multicolumn{1}{c}{$a\times b$ [95\% CI]} \\
\midrule
Words & $+0.291^{***}$ & $+0.31^{***}$ & $+0.0910^{\dagger}$ \footnotesize{[+0.0860, +0.0963]} & $+0.29^{***}$ & $+0.0850^{\dagger}$ \footnotesize{[+0.0803, +0.0898]} \\
Bulleted lines & $+0.161^{***}$ & $-0.04^{***}$ & $-0.0066^{\dagger}$ \footnotesize{[-0.0077, -0.0054]} & $-0.03^{***}$ & $-0.0056^{\dagger}$ \footnotesize{[-0.0067, -0.0045]} \\
Emoji & $+0.083^{***}$ & $-0.01^{*}$ & $-0.0005^{\dagger}$ \footnotesize{[-0.0009, -0.0001]} & $0.00$ & $+0.0001$ \footnotesize{[-0.0003, +0.0005]} \\
Requests to comment & $+0.009^{***}$ & $+0.21^{***}$ & $+0.0020^{\dagger}$ \footnotesize{[+0.0011, +0.0030]} & $+0.47^{***}$ & $+0.0044^{\dagger}$ \footnotesize{[+0.0023, +0.0066]} \\
Requests to share & $+0.009^{**}$ & $+0.09^{***}$ & $+0.0008^{\dagger}$ \footnotesize{[+0.0003, +0.0014]} & $+0.10^{***}$ & $+0.0008^{\dagger}$ \footnotesize{[+0.0003, +0.0014]} \\
Requests to follow & $+0.012^{***}$ & $+0.09^{***}$ & $+0.0010^{\dagger}$ \footnotesize{[+0.0005, +0.0016]} & $+0.12^{***}$ & $+0.0014^{\dagger}$ \footnotesize{[+0.0008, +0.0022]} \\
Ending with a question & $-0.059^{***}$ & $-0.06^{***}$ & $+0.0034^{\dagger}$ \footnotesize{[+0.0026, +0.0041]} & $-0.01^{*}$ & $+0.0007^{\dagger}$ \footnotesize{[0.0000, +0.0013]} \\
Hashtags & $-0.057^{***}$ & $-0.04^{***}$ & $+0.0022^{\dagger}$ \footnotesize{[+0.0015, +0.0029]} & $-0.08^{***}$ & $+0.0044^{\dagger}$ \footnotesize{[+0.0033, +0.0055]} \\
Links & $-0.275^{***}$ & $-0.19^{***}$ & $+0.0517^{\dagger}$ \footnotesize{[+0.0480, +0.0549]} & $-0.18^{***}$ & $+0.0483^{\dagger}$ \footnotesize{[+0.0446, +0.0518]} \\
Sentiment & $-0.015^{***}$ & $+0.30^{***}$ & $-0.0046^{\dagger}$ \footnotesize{[-0.0071, -0.0021]} & $+0.18^{***}$ & $-0.0027^{\dagger}$ \footnotesize{[-0.0042, -0.0012]} \\
\midrule
Indirect, post length & & $59.9\%$ & $+0.0910^{\dagger}$ \footnotesize{[+0.0860, +0.0963]} & $44.9\%$ & $+0.0850^{\dagger}$ \footnotesize{[+0.0803, +0.0898]} \\
Indirect, links & & $34.0\%$ & $+0.0517^{\dagger}$ \footnotesize{[+0.0480, +0.0549]} & $25.6\%$ & $+0.0483^{\dagger}$ \footnotesize{[+0.0446, +0.0518]} \\
Indirect, the four requests & & $4.8\%$ & $+0.0072^{\dagger}$ \footnotesize{[+0.0057, +0.0088]} & $3.9\%$ & $+0.0073^{\dagger}$ \footnotesize{[+0.0048, +0.0098]} \\
Indirect, the three formatting features & & $-3.2\%$ & $-0.0048^{\dagger}$ \footnotesize{[-0.0063, -0.0035]} & $-0.6\%$ & $-0.0011$ \footnotesize{[-0.0028, +0.0004]} \\
Indirect, sentiment & & $-3.1\%$ & $-0.0046^{\dagger}$ \footnotesize{[-0.0071, -0.0021]} & $-1.4\%$ & $-0.0027^{\dagger}$ \footnotesize{[-0.0042, -0.0012]} \\
Indirect, all ten features & &  & $+0.1404^{\dagger}$ \footnotesize{[+0.1345, +0.1467]} &  & $+0.1368^{\dagger}$ \footnotesize{[+0.1309, +0.1429]} \\
Direct effect of the AI score ($c'$) & &  & $+0.012$ \footnotesize{(0.013)} &  & $+0.053^{***}$ \footnotesize{(0.014)} \\
Total effect of the AI score ($c$) & &  & $+0.152^{***}$ \footnotesize{(0.013)} &  & $+0.189^{***}$ \footnotesize{(0.014)} \\
Share of total through features & &  & 92.4\% &  & 72.4\% \\
\midrule
Posts / authors & \multicolumn{5}{l}{559,149 / 106,968} \\
\bottomrule
\end{tabular}}
\end{table}

\section*{Supplementary figures}

\begin{figure}[H]
\centering
\includegraphics[width=0.85\linewidth]{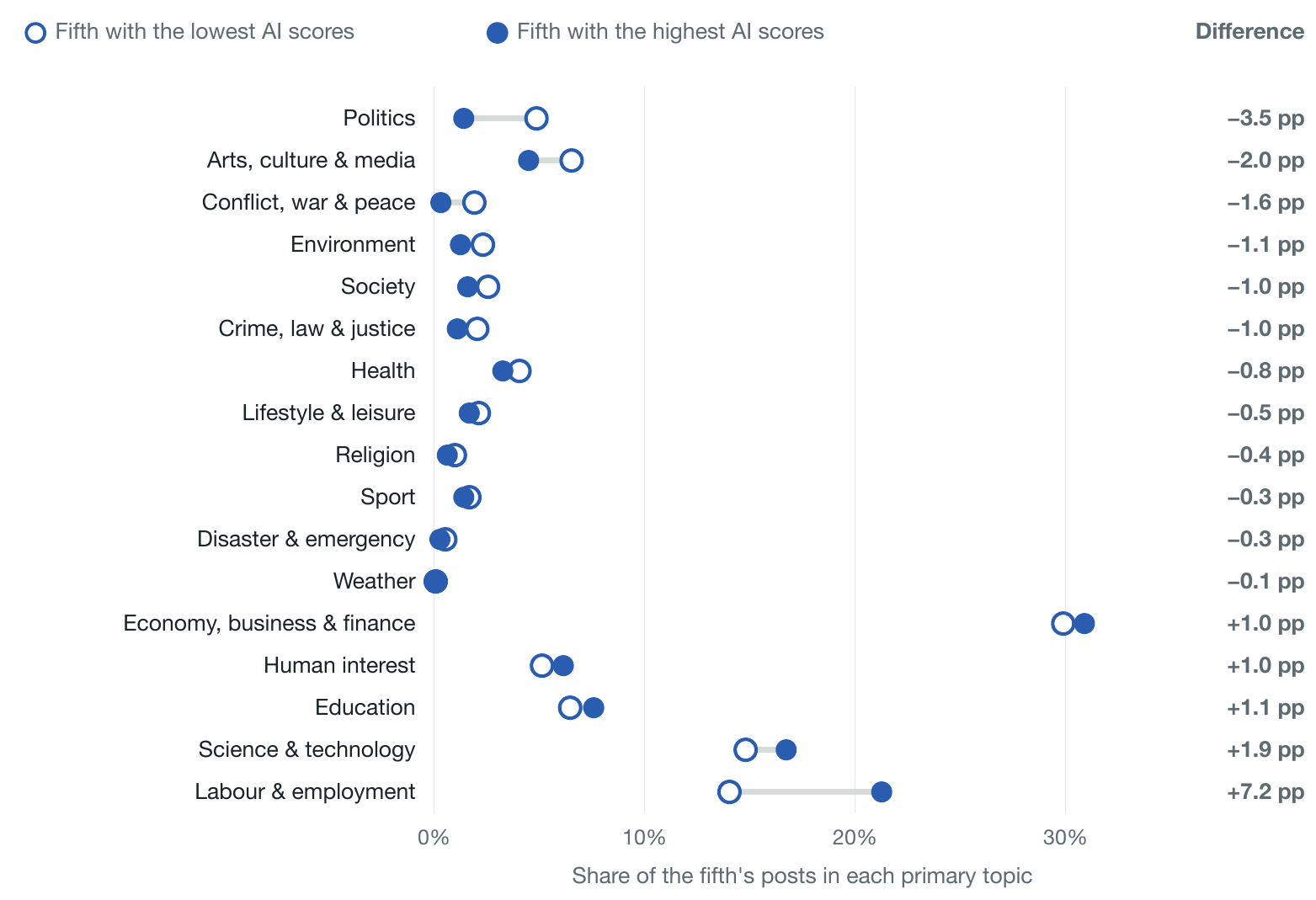}
\caption{\textbf{Topics of AI-generated posts.} Share of posts in each primary topic among the fifths of posts with the lowest (open circles) and highest (filled circles) AI scores. Values at right are the differences in percentage points. $\chi^2=12{,}773$, $P<0.001$, Cram\'er's $V=0.18$.}
\label{fig:topics}
\end{figure}

\begin{figure}[H]
\centering
\includegraphics[width=\textwidth]{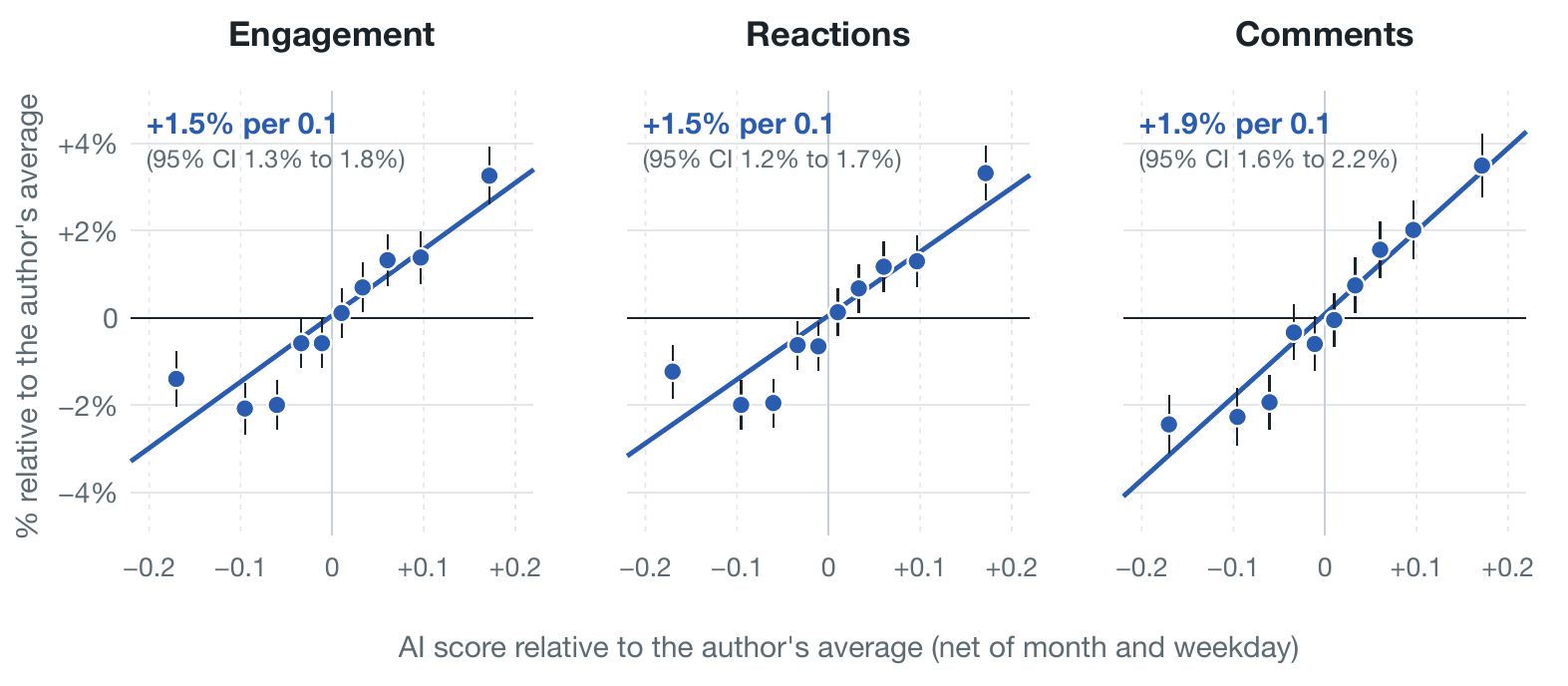}
\caption{\textbf{Engagement when a post scores above or below the author's average.} Posts were grouped into deciles of how far their AI score was from the author's average, net of month and weekday. Points are each decile's mean engagement, reactions, and comments relative to the author's average, with 95\% CIs (standard errors clustered by author). Lines are the within-author regression slopes, and the number in each panel is the change per 0.1 increase in the AI score. Data are from 559,149 posts by 106,968 authors.}
\label{fig:engagement_bins}
\end{figure}

\begin{figure}[H]
\centering
\begin{minipage}[t]{0.49\linewidth}
    \centering
    \includegraphics[width=\linewidth]{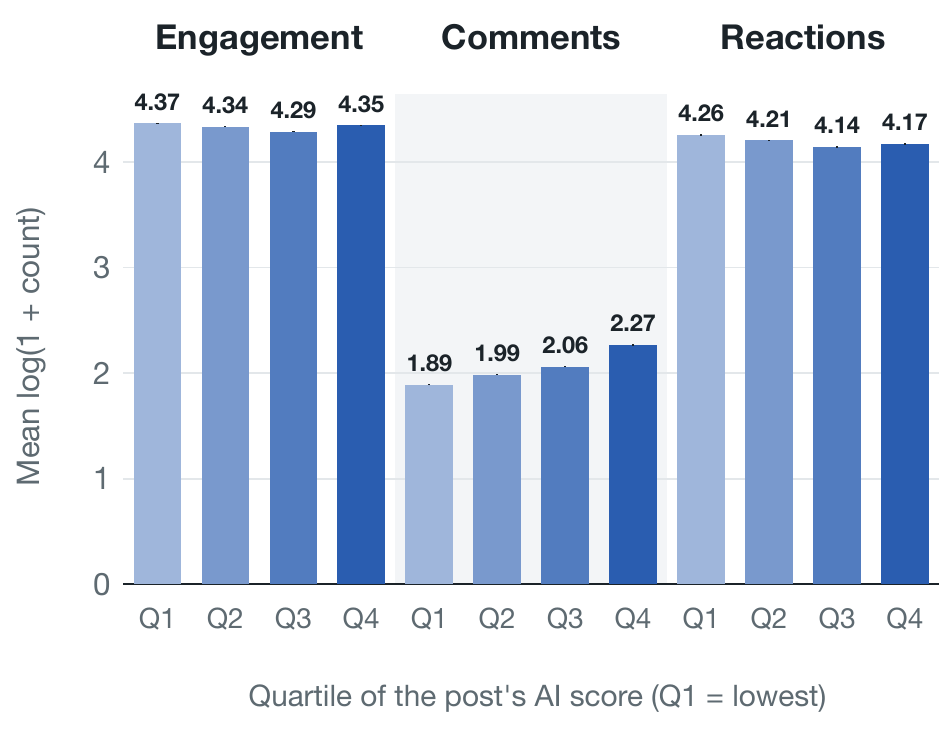}\\[2pt]
    {\small (A)}
\end{minipage}
\hfill
\begin{minipage}[t]{0.49\linewidth}
    \centering
    \includegraphics[width=\linewidth]{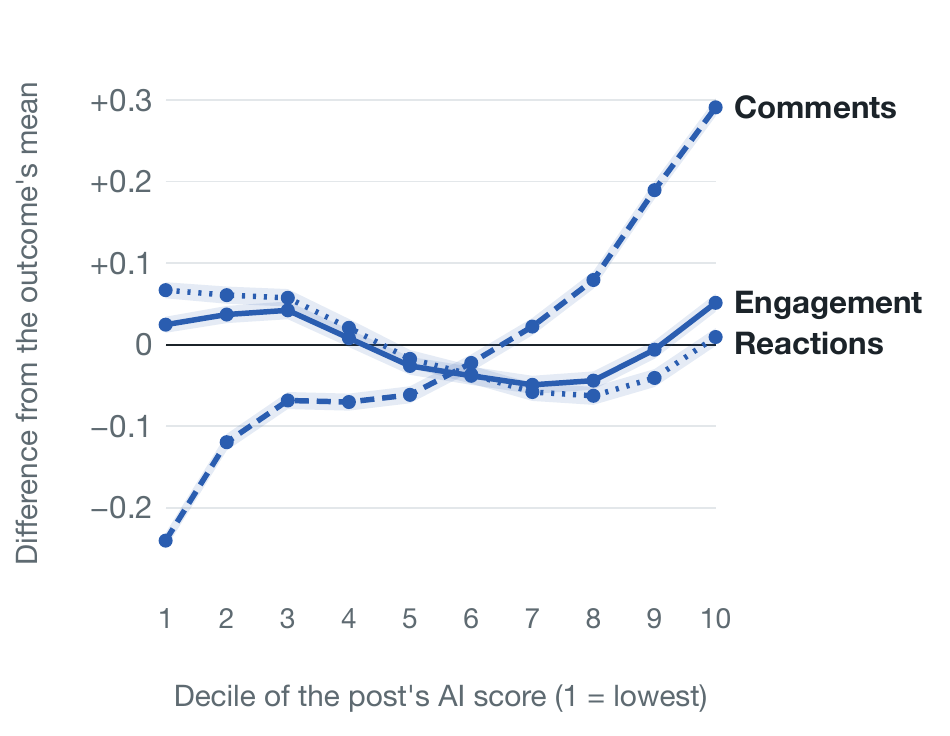}\\[2pt]
    {\small (B)}
\end{minipage}
\caption{\textbf{Engagement across levels of the AI score, across all posts.} These comparisons are across posts and not within authors. Posts are pooled across authors, so differences between authors and differences within the same author are mixed together. (\textit{A}) Mean $\log(1+\text{count})$ of total engagement, comments, and reactions by quartile of the post's AI score, with 95\% CIs. (\textit{B}) The same outcomes by decile, each centered on its sample mean, with 95\% CIs. Data are from 995,629 posts. The values are unadjusted, and the within-author estimates are in Fig.~\ref{fig:engagement} and Fig.~\ref{fig:engagement_bins}.}
\label{fig:engagement_raw}
\end{figure}

\begin{figure}[H]
\centering
\includegraphics[width=0.5\linewidth]{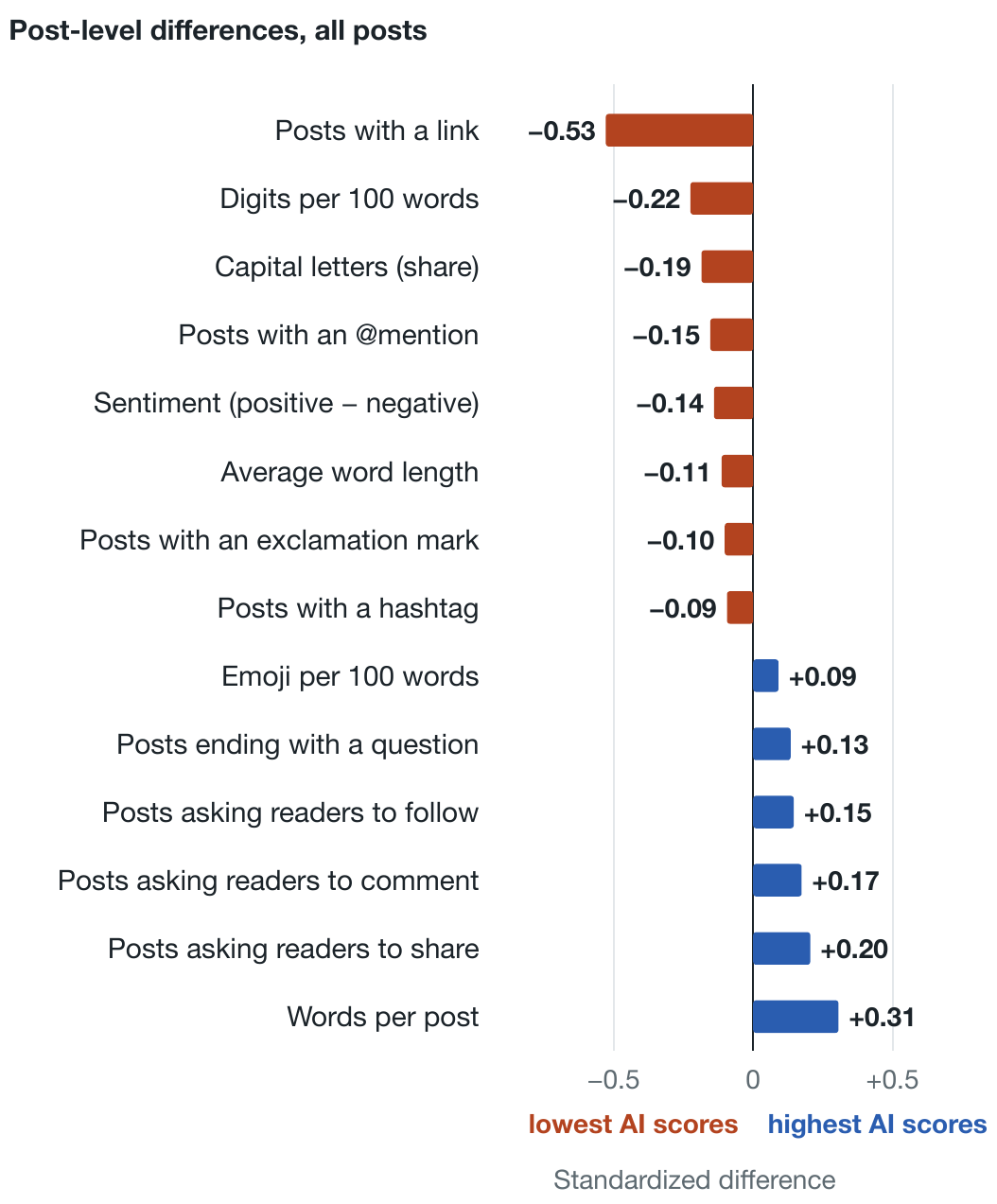}
\caption{\textbf{Post-level differences, all posts.} Standardized differences between the fifths of all 995,629 posts with the highest and lowest AI scores (199,126 posts per fifth). Posts are pooled across authors, so differences between authors and differences within the same author are mixed together. Blue and orange bars are as in Fig.~\ref{fig:authors}\textit{A}. All differences are significant at $P<0.001$ (Welch $t$ tests). The characteristics are defined in Table~\ref{tab:traits_def}.}
\label{fig:post_characteristics}
\end{figure}

\ifSubfilesClassLoaded{\bibliography{references}}{}

\end{document}

\end{document}